\documentclass[preprint,12pt]{elsarticle}
\usepackage[pdfborder={0 0 0}]{hyperref}
\usepackage{amsthm,amsmath,amsfonts,latexsym,amssymb}
\usepackage{graphics,fancyhdr,graphicx}
\usepackage[ruled,linesnumbered]{algorithm2e}
\usepackage[margin=1in]{geometry}
\usepackage{multirow,booktabs}
\usepackage[table]{xcolor}
\usepackage{listings}
\usepackage{tabularx}
\usepackage{placeins}
\usepackage{comment}
\usepackage{lscape}
\usepackage{bm}
\usepackage{hyperref}
\usepackage{cleveref}
\usepackage{mathrsfs}
\usepackage{arydshln}
\usepackage{appendix}
\usepackage{soul}
\definecolor{custom-blue}{RGB}{3,69,173}
\hypersetup{
    colorlinks = true,
    urlcolor   = blue,
    citecolor  = custom-blue,
}
\biboptions{numbers,sort&compress}
\definecolor{listinggray}{gray}{0.9}
\definecolor{lbcolor}{rgb}{0.9,0.9,0.9}
\definecolor{Darkgreen}{RGB}{0,100,0}
\usepackage[font=small,labelfont=bf, textfont=it]{caption}

\usepackage{float}
\usepackage{booktabs}
\usepackage{array}
\usepackage{multirow}
\usepackage{tabularray}
\usepackage{hhline}
\usepackage{algpseudocode}
\usepackage{mathtools}
\usepackage{subcaption}
\usepackage{enumitem}
\usepackage{lipsum}
\usepackage[most]{tcolorbox}

\makeatletter
\def\ps@pprintTitle{%
 \let\@oddhead\@empty
 \let\@evenhead\@empty
 \def\@oddfoot{}%
 \let\@evenfoot\@oddfoot}
\makeatother

\newtcolorbox{textbox}[1]{
    title={\hspace*{0mm}#1},
}

\newtcolorbox{response}[1]{
    colframe=custom-blue,
    colback=black!8,
    title={\hspace*{0mm}#1},
}

\begin{document}
\abovedisplayskip=6.0pt
\belowdisplayskip=6.0pt


\newcommand{\Review}[2][blue]{{\textcolor{#1}{#2}}}

\begin{frontmatter}

\title{
Agentic Risk-Aware Set-Based Engineering Design
}

\author[1]{Varun Kumar}
\ead{varun_kumar2@brown.edu}
\author[2]{George Em Karniadakis\corref{cor1}}
\ead{george_karniadakis@brown.edu}

\address[1]{School of Engineering, Brown University}
\address[2]{Division of Applied Mathematics, Brown University}

\cortext[cor1]{Corresponding author.}

\begin{abstract}
\noindent
This paper introduces a multi-agent framework guided by Large Language Models (LLMs) to assist in the early stages of engineering design, a phase often characterized by vast parameter spaces and inherent uncertainty. Operating under a human-in-the-loop paradigm and demonstrated on the canonical problem of aerodynamic airfoil design, the framework employs a team of specialized agents: a Coding Assistant, a Design Agent, a Systems Engineering Agent, and an Analyst Agent - all coordinated by a human Manager. Integrated within a set-based design philosophy, the process begins with a collaborative phase where the Manager and Coding Assistant develop a suite of validated tools, after which the agents execute a structured workflow to systematically explore and prune a large set of initial design candidates. A key contribution of this work is the explicit integration of formal risk management, employing the Conditional Value-at-Risk (CVaR) as a quantitative metric to filter designs that exhibit a high probability of failing to meet performance requirements, specifically the target coefficient of lift ($C_L$). The framework automates labor-intensive initial exploration through a global sensitivity analysis conducted by the Analyst agent, which generates actionable heuristics to guide the other agents. The process culminates by presenting the human Manager with a curated final set of promising design candidates, augmented with high-fidelity Computational Fluid Dynamics (CFD) simulations. This approach effectively leverages AI to handle high-volume analytical tasks, thereby enhancing the decision-making capability of the human expert in selecting the final, risk-assessed design.
\end{abstract}
 
\end{frontmatter}

\section{Introduction}
\label{sec:intro}
Engineering design is a profoundly complex and multifaceted process, characterized by a non-linear workflow and frequent iterative cycles. The conceptual design phase, in particular, represents the most critical juncture in product development, where decisions can commit up to 75\% of the total product lifecycle cost \cite{design_process_Ullman}. During this phase, designers must navigate a vast and often poorly defined design space to identify promising solutions. This initial exploration is inherently iterative, as a wide array of alternatives must be considered and progressively refined. The conventional approach to this challenge, often termed point-based design, involves an early commitment to a single design concept, which is then subjected to successive refinements and optimizations. While straightforward, this method introduces significant inflexibility and risk. Committing to a single solution when the problem is ill-defined can lead to design fixation, a well-documented cognitive bias that impedes the exploration of alternative, and potentially superior, solutions \cite{design_fixation}. The cascading effects of such an early decision can result in costly, late-stage redesigns when unforeseen constraints or better alternatives emerge.

To address the profound limitations of point-based methods, the Set-Based Design (SBD) paradigm, which has its roots in the highly successful Toyota Production System, offers a more robust and flexible alternative \cite{sbd_toyota}. Rather than selecting a single point solution and refining it, SBD advocates for the simultaneous exploration of broad sets of design alternatives. The design team works concurrently with multiple potential solutions, maintaining ambiguity and delaying critical decisions until more information is available. This process involves gradually narrowing the design space by eliminating entire regions that are proven to be infeasible or non-promising, thereby mitigating the risk of premature commitment and fostering a more comprehensive exploration of potential solutions \cite{sbd_what_is_it}. Mathematically, this convergent process can be represented as a sequence of operations on sets of design parameters. If we define the initial, continuous design space as \( \mathcal{D} \subseteq \mathbb{R}^n \), the SBD process can be modeled as a sequence of discrete sets \( S_k \) at iteration \( k \), such that \( \mathcal{D} \supseteq S_0 \supseteq S_1 \supseteq \dots \supseteq S_f \), where \( S_f \) is the final set of selected designs. Each step in this sequence is governed by the relation \( S_{k+1} = \mathcal{F}_k (\mathcal{M}_k(S_k)) \), where \( \mathcal{M}_k \) is a mapping operator representing design evaluation or modification, and \( \mathcal{F}_k \) is a filtering operator that narrows the set based on performance, feasibility, or risk criteria.

An effective and intelligent design filtering strategy is the cornerstone of the SBD methodology. One powerful approach for implementing the filtering operator \( \mathcal{F}_k \) is risk-based design selection. In this context, risk is formally defined as a function of both the probability and the consequence of failure to meet a performance objective \cite{risk_quantitative}, and it explicitly accounts for various sources of uncertainty. These uncertainties are typically categorized as either aleatory (inherent randomness in the system or environment) or epistemic (a reducible lack of knowledge or inaccuracy in predictive models) \cite{bayesian_kennedy}. By evaluating design candidates based on their expected risk profile, this method ensures that the selected designs are robust to variations in operating conditions and model-form uncertainty. The principles of incorporating risk into design decisions are well-established, drawing from decision theory and robust design methodologies \cite{engg_decisions_hazelrigg, evidence_theory_design}. To manage computational costs in early-stage design exploration, performance evaluations typically rely on simplified, low-fidelity physical models (e.g., potential flow codes in aerodynamics) \cite{approximation_alexandrov}. While these models offer rapid assessments, their inherent simplifications and missing physics-such as neglecting viscous effects or turbulence-can lead to grossly inaccurate performance predictions, potentially causing superior designs to be prematurely discarded.

To bridge this gap between computational cost and predictive accuracy, surrogate models, or metamodels, have emerged as an indispensable tool in modern engineering design \cite{metamodels}. Among the various types of surrogates, those based on neural networks (NNs) have shown significant promise in recent years for their ability to approximate highly non-linear, high-dimensional functions. These models are trained on a limited set of high-fidelity data points and learn the complex mapping between design parameters and performance outcomes. Their primary advantage is a significant shift in computational timescale, reducing evaluation times from hours or days for a single high-fidelity simulation to milliseconds. This speed enables the exhaustive evaluation of thousands of design candidates, making large-scale SBD exploration computationally feasible. Furthermore, unlike simplified physics models, well-trained NNs can capture the intricate physical phenomena present in the high-fidelity data, leading to far more reliable evaluations within the SBD filtering process.

This paper addresses the existing research gap in the integration and intelligent orchestration of these advanced design paradigms. While SBD, risk analysis, and surrogates are powerful individually, their synergistic combination into a cohesive, semi-autonomous workflow remains a significant challenge. We introduce a novel framework that integrates these concepts for the conceptual design of airfoils, orchestrated by an innovative multi-agent system driven by Large Language Models (LLMs). This framework leverages the reasoning and coordination capabilities of LLMs to manage the iterative design loop of evaluation, filtering, and review. The primary contributions of this work are threefold: (1) the development and application of a comprehensive set-based design methodology for airfoil design, featuring a robust, risk-based filtering mechanism; (2) the implementation of a multi-stage design evaluation and refinement process powered by pre-trained neural network surrogates for high-speed, high-fidelity assessment; and (3) the introduction of a novel LLM-driven multi-agent framework for automating design evaluation and ranking, augmented by a human-in-the-loop feedback system for final design validation. The remainder of this paper details the architecture of this framework, demonstrates its application to an airfoil design case study, and discusses the results and implications for future intelligent design systems. The complete workflow for this framework is shown in Figure \ref{fig:workflow}. Additional background review on this topic is presented in Appendix \ref{sec:related_works}.

\begin{figure}[h]
    \centering
    \includegraphics[width=1\linewidth]{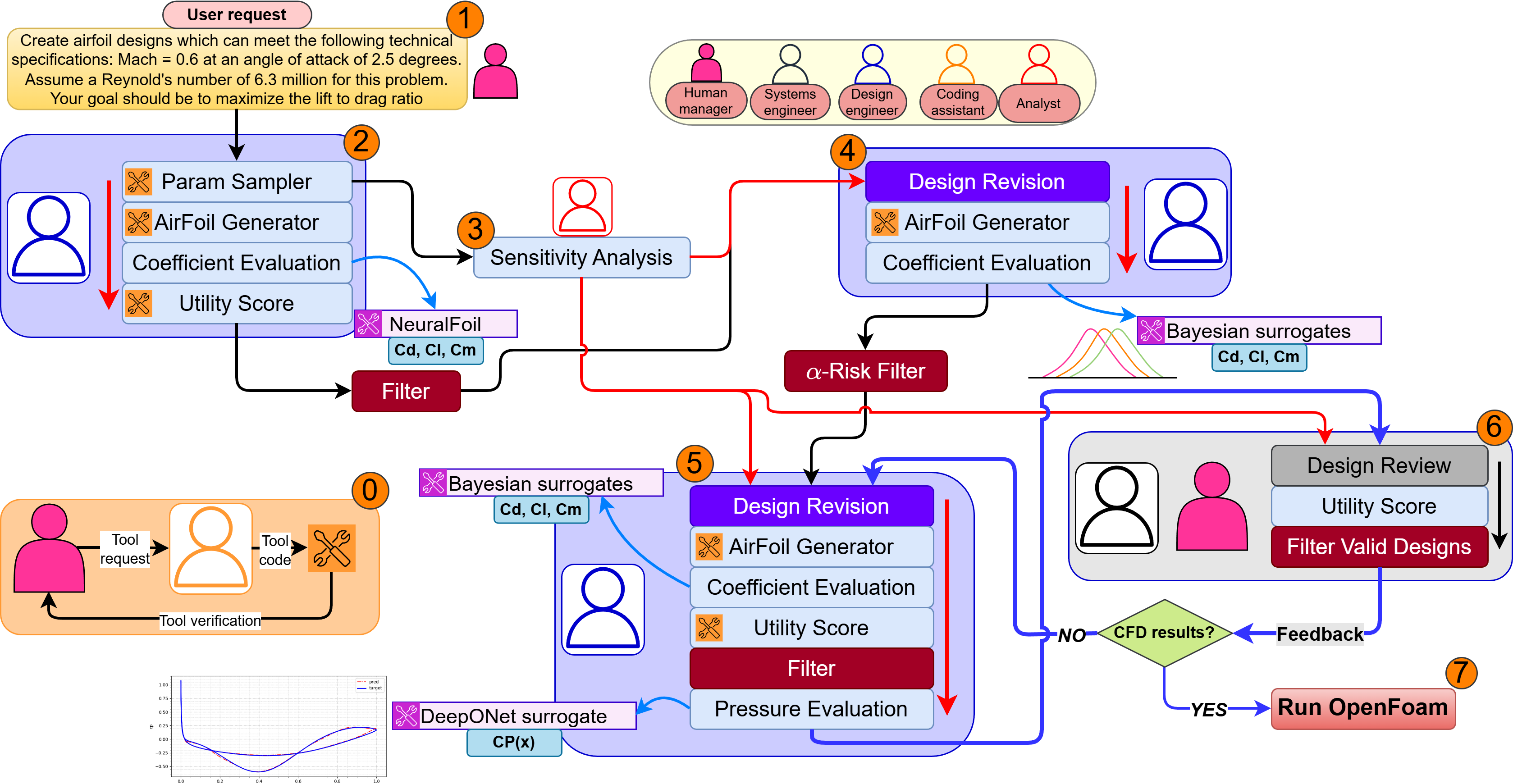}
    \caption{Schematic for LLM-based set-based design workflow with risk filtering strategy. The LLM workflow consists of four agents: \emph{Design Engineer}, \emph{Systems Engineer}, \emph{Coding Assistant}, and \emph{Analyst}, each assigned a designated role and task. The workflow utilizes different tools that are developed by the Coding Assistant following human manager's instructions. The workflow starts with the human manager providing a design problem to the Design Engineer. The Design Engineer, equipped with a set of engineering tools, then proceeds to developing, analyzing, and filtering design candidates based on certain rules. After a sequence of design filtering and improvements, an iterative design review and improvement cycle commences with the Systems Engineer reviewing and rating the design solutions based on a utility score, in comparison with a benchmark airfoil design. Final design candidates are reviewed jointly by the Systems Engineer and human manager, who determines which designs to retain in the design process. The process ends once the human manager determines a set of viable candidate designs and requests for CFD analysis of these candidates.}
    \label{fig:workflow}
\end{figure}

\section{Problem statement}
\label{sec:problem_statement}
The design of an airfoil profile is a classic engineering optimization challenge, involving a delicate trade-off between maximizing lift and minimizing drag under specific flight conditions. Key aerodynamic metrics, such as the coefficients of lift ($C_L$), drag ($C_D$), and pitching moment ($C_M$), are highly sensitive to the airfoil's geometry and the flow regime, which is characterized by the Reynolds number (Re), Mach number (Ma), and angle of attack (AoA). The primary goal of airfoil design problem is to identify a geometry that yields superior performance for a given set of operating conditions.

To facilitate an automated design process, the airfoil geometry must be described by a finite set of parameters that can be manipulated during iterative design process. A widely adopted method for this purpose is the Class-Shape Transformation (CST) parameterization, introduced by Kulfan \cite{kulfan}. Unlike other airfoil parameterization methods such as NACA, the CST method does not require well-defined equations for representing airfoil profile. It also provides a robust and intuitive way to represent a wide range of airfoil shapes with a relatively small number of design variables. The vertical coordinates of the upper ($y_u$) and lower ($y_l$) airfoil surfaces are defined as a product of a class function, $C(x)$, and a shape function, $S(x)$, plus a term for a finite trailing edge thickness, $y_{TE}$. The formulatio1n in a normalized coordinate system ($x \in [0, 1]$) is given by:

\begin{equation}
    y(x) = C(x)S(x) + x \cdot y_{TE} \, .
\end{equation}

The class function, $C_{N_2}^{N_1}(x) = x^{N_1}(1-x)^{N_2}$, defines the fundamental topology of the airfoil class. For conventional airfoils, the exponents $N_1=0.5$ and $N_2=1.0$ are chosen to ensure a rounded leading edge and a sharp, finite-angle trailing edge, respectively. The shape function is a linear combination of Bernstein polynomials, which provides the detailed contour:

\begin{equation}
    S(x) = \sum_{i=0}^{n} w_i B_{i,n}(x) = \sum_{i=0}^{n} w_i \binom{n}{i} x^i (1-x)^{n-i} \,. \label{eqn:cst_eqn}
\end{equation}

where $n$ is the number of weights defining the airfoil surface. In this formulation, the weights, $w_i$, associated with each Bernstein polynomial, $B_{i,n}(x)$, serve as the design variables that control the airfoil's final shape. A distinct set of weights is used for the upper and lower surfaces, providing comprehensive control over the airfoil's camber and thickness distribution. In this study, we adopt the range of CST parameters as defined in \cite{airfoil_data} which allows us to use this existing CFD data for training our neural surrogates.

This study frames the challenge as a single-point design problem. The primary objective is to design an efficient airfoil that maximizes the lift coefficient ($C_L$) under specific flow conditions. The chosen operating point is defined by a Reynolds number of Re = 6.3 million, a freestream Mach number of Ma = 0.6, and an angle of attack of AoA = 2.5 degrees. These conditions were specifically selected because they correspond to the well-documented experimental test case for the RAE2822 airfoil \cite{rae2822_report}. The availability of extensive experimental and computational data for this case provides a crucial benchmark for validating the simulation results and assessing the performance of the final design generated by our LLM-agent framework. While the primary objective is lift maximization, other critical performance metrics, including the drag coefficient ($C_D$), the moment coefficient ($C_M$), and the surface pressure distribution ($C_P$), will be evaluated for airfoil designs to ensure a well-rounded and aerodynamically efficient design.

\section{Agentic Design as Sequential Decision-Making Under Uncertainty}

In this section, we present a theoretical interpretation of the proposed multi-agent design framework, formalizing the agentic workflow as a sequential decision-making process under uncertainty. Specifically, we cast the methodology as an iterative procedure over sets of candidate designs, governed by stochastic evaluation operators and risk-aware filtering mechanisms. This perspective provides a mathematical foundation for understanding the roles of individual Large Language Model (LLM) agents, the treatment of epistemic and aleatory uncertainties, and the incorporation of quantitative risk metrics to guide the efficient exploration of high-dimensional engineering design spaces.

We adopt the set-based design (SBD) paradigm \cite{set_based_ward}, which shifts the focus from optimizing a single point to progressively narrowing a space of possibilities. Let $\mathcal{D} \subset \mathbb{R}^n$ denote the continuous design space parameterized by $n$ variables, such as the Class Shape Transformation (CST) coefficients used to define aerodynamic profiles. The SBD process can be mathematically represented as a monotonically decreasing sequence of subsets:
\begin{equation}
\mathcal{D} \supseteq S_0 \supseteq S_1 \supseteq \cdots \supseteq S_K,
\end{equation}
where $S_k$ denotes the active set of candidate designs at the $k$-th iteration. The evolution of this candidate set is driven by the application of two distinct, sequentially applied operators. Formally, the transition from $S_k$ to $S_{k+1}$ is expressed as:
\begin{equation}
S_{k+1} = \mathcal{F}_k\big( \mathcal{M}_k(S_k) \big),
\end{equation}
where $\mathcal{M}_k: S_k \rightarrow \tilde{S}_k$ represents a modification operator responsible for the generation and refinement of candidate geometries, effectively exploring the local design space. Conversely, $\mathcal{F}_k: \tilde{S}_k \rightarrow S_{k+1}$ acts as a filtering operator that discards inferior designs based on performance and risk criteria. Within our framework, these abstract operators are instantiated by specialized LLM agents: the Design Agent executes the modification mapping $\mathcal{M}_k$, while the Systems Engineering Agent, guided by risk-aware filters, implements the evaluation mapping $\mathcal{F}_k$.

A critical aspect of early-stage engineering design is the presence of both epistemic uncertainty (e.g., surrogate model inaccuracies) and aleatory uncertainty (e.g., variable operating conditions) \cite{uncertainty_oberkampf}. Consequently, each design vector $x \in \mathcal{D}$ is evaluated through a stochastic performance model. We denote the performance metrics of a design, such as the coefficients of lift ($C_L$), drag ($C_D$), and moment ($C_M$), as a random vector $Y(x)$ distributed according to a probability density function $p(y \mid x)$. This distribution is typically induced by the underlying predictive tools, such as Bayesian neural networks or other probabilistic surrogate models. To aggregate these multivariate metrics into a unified evaluation criterion, we define a scalar utility functional \cite{utility_theory_book}:
\begin{equation}
U(x) = \mathbb{E}[u(Y(x))],
\end{equation}
where the function $u(\cdot)$ maps the stochastic performance vector to a deterministic utility score, reflecting the human Manager's design objectives.

Basing decisions solely on expected utility can leave the design vulnerable to critical failures, particularly when the performance distribution exhibits heavy tails \cite{risk_design_Rockafellar}. To ensure robustness under uncertainty, the filtering operator $\mathcal{F}_k$ explicitly incorporates formal risk measures. For a given performance metric of interest $X$ (e.g., $C_L$), we utilize the Conditional Value-at-Risk (CVaR) \cite{cvar_optimization} at a specified confidence level $\alpha \in (0,1)$. The CVaR represents the expected value of the worst-case outcomes and is defined as:
\begin{equation}
\text{CVaR}_\alpha(X) = \mathbb{E}\big[X \mid X \geq \text{VaR}_\alpha(X)\big],
\end{equation}
where $\text{VaR}_\alpha$ is the corresponding $\alpha$-quantile of the distribution. The filtering operator is subsequently defined as an indicator function:
\begin{equation}
\mathcal{F}_k(x) = 
\begin{cases} 
1, & \text{if } 
    \begin{cases} 
    U(x) \geq \tau^*, & k \in \mathcal{K}_U \\ 
    \text{CVaR}_\alpha(X) \geq \gamma^*, & k \in \mathcal{K}_R 
    \end{cases} \\
0, & \text{otherwise}
\end{cases}
\end{equation}
where $\tau^*$ and $\gamma^*$ are the utility and risk thresholds respectively, $\mathcal{K}_U$ and $\mathcal{K}_R$ represent the iteration using either utility function or CVaR as a risk-filtering strategy. This dual-constraint formulation guarantees that the retained designs are not only high-performing in expectation but also robust against adverse uncertainty realizations.

In this framework, the state at step $k$ is defined by the current set of candidates, $s_k = S_k$. The action $a_k = \mathcal{M}_k$ corresponds to the specific modification strategy or heuristics applied to $s_k$. The state transition is deterministically governed by the composite application of modification and filtering:
\begin{equation}
s_{k+1} = \mathcal{F}_k\big( \mathcal{M}_k(s_k) \big).
\end{equation}


Assuming mild regularity conditions on the filtering operator, the sequence of subsets $\{S_k\}$ forms a monotonically shrinking space, enforcing $S_{k+1} \subseteq S_k$. Eventually, the agentic process converges to a terminal set of designs $S_K$ that satisfies the strict criteria:
\begin{equation}
S_K = \left\{ x \in \mathcal{D} \mid U(x) \geq \tau^*, \; \text{CVaR}_\alpha(X) \geq \gamma^* \right\}.
\end{equation}
Ultimately, this theoretical formulation underscores two fundamental properties of the proposed multi-agent framework: it naturally balances exploration and exploitation through the alternation of the modification ($\mathcal{M}_k$) and filtering ($\mathcal{F}_k$) operators and it enforces structural robustness via CVaR-based risk constraints. These distinct characteristics elevate the proposed agentic framework beyond classical point-based optimization routines, establishing a basis for its application in complex, uncertain engineering design workflows and directly motivating the multi-agent architecture described in the subsequent sections.

\section{Methodology}
\label{sec:methodology}
In this section, we discuss the key elements and processes associated with our MAS framework that orchestrates the SBD paradigm.
\subsection{Agent: Coding Assistant}
A central challenge in developing robust multi-agent systems for engineering design is ensuring the agents have access to reliable tools necessary in the design workflow. To address this, our framework incorporates a specialized `Coding Assistant` powered by Gemini-2.5-pro. This agent acts as a dedicated tool-smith, translating high-level, natural language requests from a human user into executable code. This process is intentionally designed as an offline, human-in-the-loop workflow, separate from the main autonomous design loop. In this setting, the user specifies a required function: for example, a tool for the `Design Engineer' agent to generate design candidates using Latin Hypercube Sampling and the `Coding Assistant' produces the corresponding code. The human manager then evaluates this code for correctness and efficiency, enabling an iterative feedback loop for refinement until the tool is fully validated. By pre-defining and sanctioning the entire toolset before the primary design task begins, we mitigate the risks of on-the-fly code generation, such as logical errors or redundant functions. This deliberate separation of concerns supervised tool creation versus autonomous tool application is a critical architectural choice that provides the core design agents with a foundation of robust tools, thereby enhancing the overall reliability and predictability of the multi-agent system. An example prompt provided to the Coding Assistant is shown in figure \ref{fig:prompt_eg_coding_assistant}.
\begin{figure}[h]
    \centering
    \begin{textbox}{User input for creating tool: Param sampler}
    \scriptsize
    " I want to create a tool which enables the following:
    \begin{itemize}[itemsep=1pt, parsep=0pt, topsep=0pt] 
        \item Create samples from the design space for the 9 CST parameters given by
        
        <designspace>
    
            "CST1": (0.0644, 0.1932),
            
            "CST2": (0.0688, 0.2064),
            
            "CST3": (0.0961, 0.2883),
            
            "CST4": (0.0961, 0.2882),
            
            "CST5": (0.1010, 0.3030),
            
            "CST6": (0.0680, 0.2039),
            
            "CST7": (0.1126, 0.3377),
            
            "CST8": (0.0381, 0.1143),
            
            "CST9": (-0.0586, -0.0195)
            
        </designspace>
        \item Enable user to choose from different sampling strategies such as Latin Hypercube, Sobol, random etc by enabling an input argument to the tool. You can use qmc library from scipy for this.
        \item Provide an input argument for number of samples to be generated to the tool.
        \item Provide input argument to the tool for 3 scalar flow parameters: Ma (float), AoA (float), Re (float). These parameters remain the same for all samples in the design space.
        \item Create an array named `nninputs' with size: numsamples x 12, where the first three columns on the array contain the 3 flow parameter and the remaining 9 columns contain the CST parameters sampled from the design space.
        \item Create a list of `designid' such as ID-1, ID-2,....., ID-numsamples for the number of design samples.
        \item Convert array `nninputs' to a list. Then export the following dictionary as a pickle object.
        
            <exportdict>
            
                \quad `samples': nninputs (list)
                
                \quad `designid': designid
                
            </exportdict>
    \end{itemize}
  \end{textbox} 
  \caption{Sample instruction provided to the Coding Agent by the human user to create a parameter sampling tool based on specific process requirements.}
  \label{fig:prompt_eg_coding_assistant}
\end{figure}

\subsection{Agent: Design Engineer}
The Design Engineer agent serves as the primary generative and iterative agent within the multi-agent framework, embodying the role of a traditional design specialist. Its core responsibility is to navigate the design space through a  two-phase process: an initial, divergent exploration followed by a convergent,  data-driven refinement. In the initial phase, the agent autonomously generates  a diverse population of candidate airfoil geometries by sampling the  9-dimensional design space defined by the CST parameters. It then rapidly estimates key aerodynamic performance metrics, such as the coefficients of lift ($C_L$) and drag ($C_D$), using computationally inexpensive surrogate models to assign a utility score and filters a set of promising candidates. During the later convergent stages, the agent’s role shifts to focused improvement, 
operating in a collaborative feedback loop with the Systems Engineer agent and  potentially a human user. To facilitate intelligent design modifications, the  Design Engineer relies on quantitative sensitivity analysis results provided by the Analyst agent, which correlate the CST parameters to aerodynamic performance. Armed with this information, the agent can make informed,  non-random adjustments, simultaneously modifying an entire set of promising  designs to efficiently steer their performance toward a desired optimum based on system-level feedback, thereby accelerating convergence to a high-quality solution.

\subsection{Agent: Systems Engineer}
The Systems Engineer agent functions as the primary design evaluator and strategic guide within the multi-agent framework, responsible for providing assessment of airfoil designs generated by the Design Engineer. This agent uses Gemini-2.5-pro multi-modal Large Language Model, which enables a holistic analysis that mirrors the expert judgment of a human engineer. It simultaneously processes quantitative performance coefficients - specifically, the coefficients of drag ($C_D$), lift ($C_L$), and moment ($C_M$), alongside qualitative graphical data, such as surface pressure distribution curves and airfoil profiles, to identify promising design candidates. The agent operates in two distinct modes to adapt to different stages of the design workflow. In its fully autonomous mode, typically employed during the broad exploration phase, it systematically analyzes a large number of designs. It assigns a utility rating to each candidate based on predefined performance metrics and evaluates the quality of the pressure distribution by comparing it against an established benchmark, such as the RAE2822 airfoil, to autonomously filter for valid designs without human intervention. As the design process converges and the set of candidate designs is reduced to a manageable number where human involvement is pragmatic, the agent transitions to a semi-autonomous mode. In this configuration, it functions as a collaborative partner to a human manager, where both parties review and rate the final candidates in parallel. Crucially, beyond mere evaluation, the Systems Engineer agent closes the iterative design loop by providing actionable feedback. Based on its comprehensive assessment, and upon incorporating specific guidance from the human manager, it formulates and communicates specific design improvement strategies to the Design Engineer agent, thereby guiding the targeted refinement of the airfoil geometries.

\subsection{Agent: Analyst}
The Analyst serves as the primary interpretive agent within the framework, tasked with translating raw numerical data into actionable design intelligence for the Design Engineer agent. Its core function is to address a fundamental challenge associated with abstract geometric parameterizations like the Class-Shape Transformation (CST) method. While CST provides a powerful and flexible means to represent airfoils, its weight parameters ($w_i$) lack a direct, intuitive connection to classical aerodynamic shape characteristics such as maximum thickness, camber, or leading-edge radius. This decoupling makes it difficult for a designer, human or artificial, to predict how a change in a specific CST weight will affect aerodynamic performance. The Analyst agent bridges this knowledge gap by performing a quantitative sensitivity analysis, which systematically evaluates the statistical correlations between each of the CST design variables and the key performance metrics ($C_L$, $C_D$, etc.). By processing the results of numerous simulations, this agent computes a sensitivity map, often in the form of a correlation matrix, which quantifies the magnitude and direction of influence each parameter has on the design objectives. Crucially, the Analyst does not merely pass this raw data onward; its primary contribution is the synthesis of these findings into a concise set of heuristic rules and strategic recommendations. For example, it might conclude that ``increasing the third upper surface weight, $w_{u,3}$, has a strong positive correlation with the lift coefficient ($C_L$) but a weak correlation with the drag coefficient ($C_D$).'' This distilled guidance empowers the Design Engineer agent to make informed, non-random modifications, effectively steering the iterative design process toward optimal regions of the design space with greater efficiency and purpose.

\subsection{Tools}
Tools form an important component of our MAS framework, and provide our agents with the ability to execute tasks. We provide a description of all major tools including filtering strategies adopted in this section. Note that all these tools are developed by the Coding assistant on request by the human manager prior to commencement of the main design loop.

\subsubsection{Param Sampler}
The initial exploration of the design space is orchestrated by a dedicated utility, the `Param Sampler' tool, which is responsible for generating the population of candidate airfoil designs. This tool operates within a 9-dimensional design space defined by the weight parameters of the Class-Shape Transformation (CST) method. The bounds for each of the nine CST variables, which collectively describe the airfoil geometry, are precisely constrained to the ranges specified in \cite{airfoil_data}. A key feature of the sampler is its methodological flexibility, enabling the selection of various sampling strategies via an input argument. In addition to standard pseudo-random sampling, the tool integrates Quasi-Monte Carlo (QMC) methods such as Latin Hypercube and Sobol sequences, implemented using the `scipy.stats.qmc' library (this was done for consistency of Analyst response). The use of these advanced techniques is critical for ensuring a more uniform and low-discrepancy distribution of sample points across the high-dimensional parameter space, which facilitates a more efficient and comprehensive exploration compared to simple random sampling, especially when the number of evaluations is limited \cite{monte_carlo_integration}. Operationally, the tool requires the user to specify the total number of samples to generate, as well as three scalar flow parameters: Mach number (Ma), angle of attack (AoA), and Reynolds number (Re), which are extracted from the initial design request and remain constant for all generated airfoil geometries. The resulting output is a structured set of design candidates, each defined by a unique vector of nine CST parameters, ready for subsequent aerodynamic evaluation under a consistent flight regime.

\subsubsection{Airfoil Generator}
The `Airfoil Generator' tool helps in converting the abstract parametric design space into concrete airfoil geometry coordinates required for aerodynamic evaluation. This tool ingests a vector of CST weights and generates a set of cartesian coordinates that define the airfoil's profile. This transformation is governed by the CST formulation shown in equation \ref{eqn:cst_eqn}, where the vertical coordinates of the upper ($y_u$) and lower ($y_l$) surfaces are expressed as the product of a Class function, $C(x)$, and a Shape function, $S(x)$. Note that the first weight chosen on $y_l$ is equal to the first parameter of $y_u$ to maintain $\mathcal{C}^2$ continuity of the leading edge. The resulting point clouds for the upper and lower surfaces are then assembled into a single, ordered array suitable for other steps in the framework, typically by traversing the upper surface from the trailing edge to the leading edge and then back along the lower surface. 

\subsubsection{Coefficient Evaluation}
For evaluating aerodynamic performance for different designs, two different surrogate models are used. These surrogates are developed to predict aerodynamic coefficients $C_D,C_L$, and $C_M$ across a range of Ma, Re, and AoA.

\begin{itemize}
    \item \textit{NeuralFoil}: For the rapid initial screening of airfoil candidates, the Design Engineer agent utilizes NeuralFoil, a deep learning-based surrogate model developed to provide near-instantaneous predictions of aerodynamic coefficients \cite{neuralfoil}. NeuralFoil approximates the solution to the inviscid Euler equations, employing a neural network architecture to map a given airfoil geometry, defined by its coordinates or parametric representation along with the angle of attack (AoA) and Mach number (Ma) to the corresponding coefficients of lift ($C_L$), drag ($C_D$), and pitching moment ($C_M$). The validated operating range for NeuralFoil covers subsonic and transonic flow conditions, specifically for Mach numbers between 0 and 0.75 and angles of attack from -5 to 15 degrees. A critical limitation of this framework, however, is its foundation in inviscid flow theory. As such, NeuralFoil does not model viscous effects, such as boundary layer development, skin friction drag, or flow separation. Consequently, its drag predictions primarily account for wave drag and are not representative of the total drag experienced by the airfoil, and its lift predictions become unreliable near stall conditions. 
    \item \textit{Bayesian surrogate:} To facilitate a robust, risk-informed filtering process, a probabilistic modeling approach is adopted to quantify the uncertainty associated with aerodynamic performance predictions. This is achieved by developing a Bayesian Neural Network (BNN) designed to learn the complex mapping from the airfoil's geometric and operational parameters to its aerodynamic coefficients. The input vector to the model consists of the nine Class-Shape Transformation (CST) parameters defining the airfoil geometry, along with the Reynolds number (Re), Mach number (Ma), and angle of attack (AoA). Since associated uncertainties may differ for each performance metric, separate and independent BNNs are trained to predict the coefficients of drag ($C_D$), lift ($C_L$), and moment ($C_M$). The primary purpose of these networks is not to provide a single point estimate but to generate a full predictive distribution for each performance coefficient. This probabilistic output is used during the alpha-risk filtering stage, where the epistemic uncertainty of surrogate models is considered for design filtering. For details related to the Bayesian surrogate, please refer to Appendix \ref{sec:bayesian_surrogate}. 
\end{itemize}

\subsubsection{Pressure Evaluation}
\label{sec:pressure_eval_tool}
While integral aerodynamic coefficients such as lift and drag provide a primary measure of airfoil performance, a detailed analysis of the surface pressure distribution, characterized by the pressure coefficient ($C_P$) is essential for airfoil design analysis. The shape of the $C_P$ curve reveals critical flow phenomena, including the location and strength of shock waves, the onset of flow separation due to adverse pressure gradients, and the extent of laminar flow regions, all of which profoundly impact overall efficiency and operational stability. Consequently, the ability to rapidly and accurately predict the pressure distribution is important for an intelligent agentic workflow. To this end, we employ a neural operator surrogate model to learn the complex mapping from airfoil geometry to its corresponding surface pressure function. Specifically, we utilize a Deep Operator Network (DeepONet), an architecture well-suited for learning operators between infinite-dimensional function spaces \cite{DeepONet}. The DeepONet is trained to map an input function, representing the airfoil geometry via its Class Shape Transformation (CST) parameters and surface coordinates, to an output function, which is the continuous pressure coefficient distribution along the chord, $C_P(x)$. Within our multi-agent framework, this predicted $C_P$ distribution serves as a critical piece of qualitative information for the Systems Engineer agent. This agent assesses a candidate design by comparing its predicted $C_P$ curve against that of the RAE2822 airfoil, a well-established supercritical benchmark \cite{rae2822_report}. This qualitative comparison allows the agent to evaluate subjective aerodynamic features, such as the desirability of a 'rooftop' pressure distribution, and subsequently assign a rating to the design. This provides nuanced feedback that guides the design selection process beyond scalar performance metrics. Details related to our DeepONet surrogate model can be found in Appendix \ref{sec:DeepONet}.

\subsubsection{Utility Score}
A fundamental challenge in multi-objective engineering design is the comparison of candidates based on several, often conflicting, performance metrics. In airfoil design, for instance, a candidate must be evaluated on its lift ($C_L$), drag ($C_D$), and moment ($C_M$) coefficients, making direct ranking non-trivial. To address this, we employ utility theory, a formal framework for quantifying preferences under uncertainty and multiple objectives \cite{utility_theory_book}. This approach allows us to translate the vector of performance metrics for each design into a single, scalar value known as a utility score. This score represents the overall desirability of the design, enabling direct comparison and ranking. Within our multi-agent framework, this quantitative evaluation is crucial for automation. The engineering agents utilize a utility scoring tool to assess each proposed design and filter out those that do not meet a minimum performance threshold. This threshold is established by the utility score of the RAE2822 airfoil, a standard benchmark, which is calculated to be approximately 0.40 under the defined preference structure. This method provides a clear, consistent, and computationally efficient criterion for automated design selection and iteration.

To compute the overall utility score, we first define individual utility functions for each aerodynamic coefficient, which are then combined through a weighted sum that reflects their relative importance in the design objectives. The combined utility, $U_{comb}$, is expressed as:

\begin{equation}
    U_{comb} = w_{CL} U(C_L) + w_{CD} U(C_D) + w_{CM} U(C_M)
\end{equation}
where the weights are set to $w_{CL}=0.5$, $w_{CD}=0.3$, and $w_{CM}=0.2$, prioritizing lift, followed by drag and moment. The individual utility functions, $U(\cdot)$, are tailored to the specific preference for each metric. For the lift coefficient ($C_L$), where higher values are preferred, we impose a hard constraint for viability ($C_L \geq 0.5$) and model diminishing returns using a concave, square-root function. For the drag coefficient ($C_D$), where lower values are strongly preferred, an exponential decay function heavily rewards low drag and rapidly penalizes any increase. Lastly, for the pitching moment coefficient ($C_M$), considered less critical, a simple linear function maps its typical operational range to a utility score. The specific mathematical formulations for each utility function are detailed below:
\begin{align}
    U(C_L) &= \begin{cases} -5.0 & \text{if } C_L < 0.5 \\ \left( \dfrac{\text{min}\left[\text{max}(C_L, 0.5), 1.2\right] - 0.5}{1.2 - 0.5} \right)^{0.5} & \text{if } C_L \geq 0.5 \end{cases} \\
    U(C_D) &= \exp(-65 \cdot C_D) \\
    U(C_M) &= \dfrac{\text{max}\left[\text{min}(C_M,0.0), -0.30 \right] - (-0.30)}{0.30}
    \label{eqn:utility_score}
\end{align}

\subsection{Filters}
In our frameworks, filter serve the purpose of eliminating designs that do not meet a specific user criteria to narrow down the design space for the Design and Systems Engineer agents to operate on. We use two types of filters to eliminate unwanted designs. Below we explain the two types of filters used in this framework.

\subsubsection{Utility score filter}
To narrow down the design selection scope to the most promising candidates, our MAS framework employs a utility score filter as a primary mechanism for automated design down-selection. This filter is integrated into the agents' toolbox and is systematically applied each time a new set of designs is generated. The filtering strategy adapts to the design stage, reflecting a progression from broad exploration to focused refinement. In the initial phase, the filter relies on performance coefficients predicted by a low-fidelity, rapid surrogate model such as NeuralFoil. At this stage, the objective is to quickly prune the design space by eliminating any candidates that fail to meet a minimum performance benchmark. This is achieved by discarding all designs with a calculated utility score below a predefined threshold, initially set to 0.4, which corresponds to the performance of the baseline RAE2822 airfoil. As the design process advances into subsequent iterative loops, the evaluation becomes more stringent. The system transitions to using the higher-fidelity Bayesian surrogate model for more accurate coefficient estimation. Concurrently, the filtering criteria are enhanced to incorporate not only the quantitative utility score but also a qualitative assessment of the airfoil's pressure distribution ($C_P$). In these later stages, the agents use the filter to eliminate designs that do not meet both the utility score target and the desired aerodynamic characteristics of the pressure plot when compared against the established benchmark. This dual-criteria approach ensures that surviving designs demonstrate good performance, both in terms of integral coefficients and physically realistic aerodynamic behavior.

\subsubsection{$\alpha$-risk filter}
In a set-based design paradigm, where numerous candidate designs are evaluated concurrently, effective filtering mechanisms are essential for focusing effort on the most promising options. While the utility score provides a measure of expected performance, it does not fully account for the risk stemming from uncertainties, which can be aleatoric (inherent randomness) or epistemic (lack of knowledge), the latter being particularly relevant when using predictions from surrogate models. To ensure the selection of robust designs that are resilient to these uncertainties, a risk-based filtering approach is employed as a secondary check. This method goes beyond mean performance to assess the potential for underperformance. For this purpose, we adopt the Conditional Value at Risk (CVaR) as a coherent risk metric \cite{cvar_optimization}. CVaR measures the expected value of a performance metric given that it falls below a certain quantile of its distribution. 

For a given design and its associated lift coefficient ($C_L$) represented by a random variable, the CVaR at a confidence level $\alpha$ is defined as the conditional expectation of $C_L$ below the Value at Risk (VaR), which is the $(1-\alpha)$-quantile of its distribution. Mathematically, for a random variable $X$, $CVaR_{\alpha}(X) = \mathbb{E}[X | X \geq VaR_{\alpha}(X)]$. We estimate CVaR using the historical method, which leverages the probabilistic nature of the Bayesian surrogate by generating a large number of independent samples from its posterior predictive distribution for each design. The CVaR is then calculated as the empirical mean of the worst $(1-\alpha)\%$ of these samples. This metric is particularly advantageous for design assessment because, unlike VaR, it captures the magnitude of potential underperformance in the tail of the distribution, not just the probability of a shortfall. This allows the agents to filter out designs that, despite having a favorable expected utility, carry an unacceptably high risk of failing to meet a minimum performance threshold, $VaR_{CL\_target}$. Note that since our goal is to identify designs that are greater than a minimum $C_L$, we invert the sign of $VaR_{\alpha}(CL)$. The implementation of this filtering step is detailed in Algorithm \ref{alg:cvar}.
\vspace{5pt}

\begin{algorithm}[h]
\DontPrintSemicolon
  
  \KwIn{Set of $N$ designs $\{d_i\}_{i=1}^N$, Bayesian surrogate model $M$, number of samples $m$, confidence level $\alpha_{CL}$, performance threshold $VaR_{CL\_target} = -0.70$}
  \KwOut{A filtered set of robust designs $D_{robust}$}
  \Begin{
    \text{Initialize}:  $D_{robust} \leftarrow \emptyset$ \;
    \For{$d_i \in \{d_i\}_{i=1}^N$}{
        Generate $m$ i.i.d. samples of lift coefficient $\{C_{L,i}^{(j)}\}_{j=1}^m$ using the posterior predictive distribution from surrogate $M$ for design $d_i$ \;
        Sort the samples in ascending order: $C_{L,i}^{(1)'} \leq C_{L,i}^{(2)'} \leq ... \leq C_{L,i}^{(m)'}$ \;
        Determine the number of tail samples, $k \leftarrow \lfloor(1-\alpha_{CL}) \cdot m\rfloor$ \;
        Calculate the CVaR for the design: $CVaR_i \leftarrow \dfrac{1}{k} \sum_{j=1}^{k} C_{L,i}^{(j)'}$ \;
        \If{$CVaR_i \leq VaR_{CL\_target}$}{Add $d_i$ to $D_{robust}$}
    }
    \Return $D_{robust}$
  }
\caption{CVaR-based Filtering of Candidate Design}
\label{alg:cvar}
\end{algorithm}

\section{Workflow and Experiments}
In this section, we describe the different stages of our workflow, referenced in figure \ref{fig:workflow} and results from our experiments.

\subsection{Stage 0: Tool development}
The initial phase of our methodology, the Precursor Stage, is dedicated to the development and validation of a custom suite of software tools that form the functional core of the multi-agent system. This foundational stage follows a synergistic human-in-the-loop model, the `Manager', collaborates with the Coding Assistant. The Manager architects the entire engineering workflow, defining the precise specifications for all necessary tools, while the Coding Assistant translates these high-level instructions into executable code. The development process is inherently iterative, involving a continuous feedback loop where the Manager verifies the generated tools, often requiring two to three cycles of refinement to achieve the final, validated code. This dynamic was particularly evident in the creation of complex components, such as the neural network surrogates. While the Coding Assistant successfully generated an initial codebase, achieving optimal predictive accuracy necessitated direct intervention from the Manager, who leveraged their expertise to modify key hyperparameters and make subtle architectural adjustments. This experience underscores a key aspect of our approach: AI assistants can significantly accelerate the implementation of well-defined software, while the nuanced judgment and deep expertise of human specialists remain indispensable for complex, performance-sensitive tasks.

\subsection{Stage 1: User request}
The first stage of the workflow involves the Design agent receiving a design request from the human manager. In this study, the request to the agent is defined by a need to design airfoils with maximum lift coefficient possible at the given flow conditions. Note here that the framework supports other flow conditions within the range described in table \ref{tab:cst_space}. Also, changing design goals is possible, with modifications required in the tool set to ensure appropriate utility functions and filtering process.

\begin{figure}[H]
    \centering
    \begin{textbox}{User request for initializing workflow}
    \scriptsize
    "Create airfoil designs for the following flow conditions: Mach=0.6 at an angle of attack of 2.5 degrees. Assume a Reynold's number of 6.3 million for this problem. Your goal should be to maximize the lift coefficient of the airfoil taking into consideration other aspects such as coefficient of drag, moment coefficient, and pressure distribution over the airfoil surface"
    \end{textbox}
    \label{fig:user_req}
\end{figure}

\subsection{Stage 2: Design generation phase}
\begin{figure}[h]
    \centering
    \includegraphics[width=0.8\linewidth]{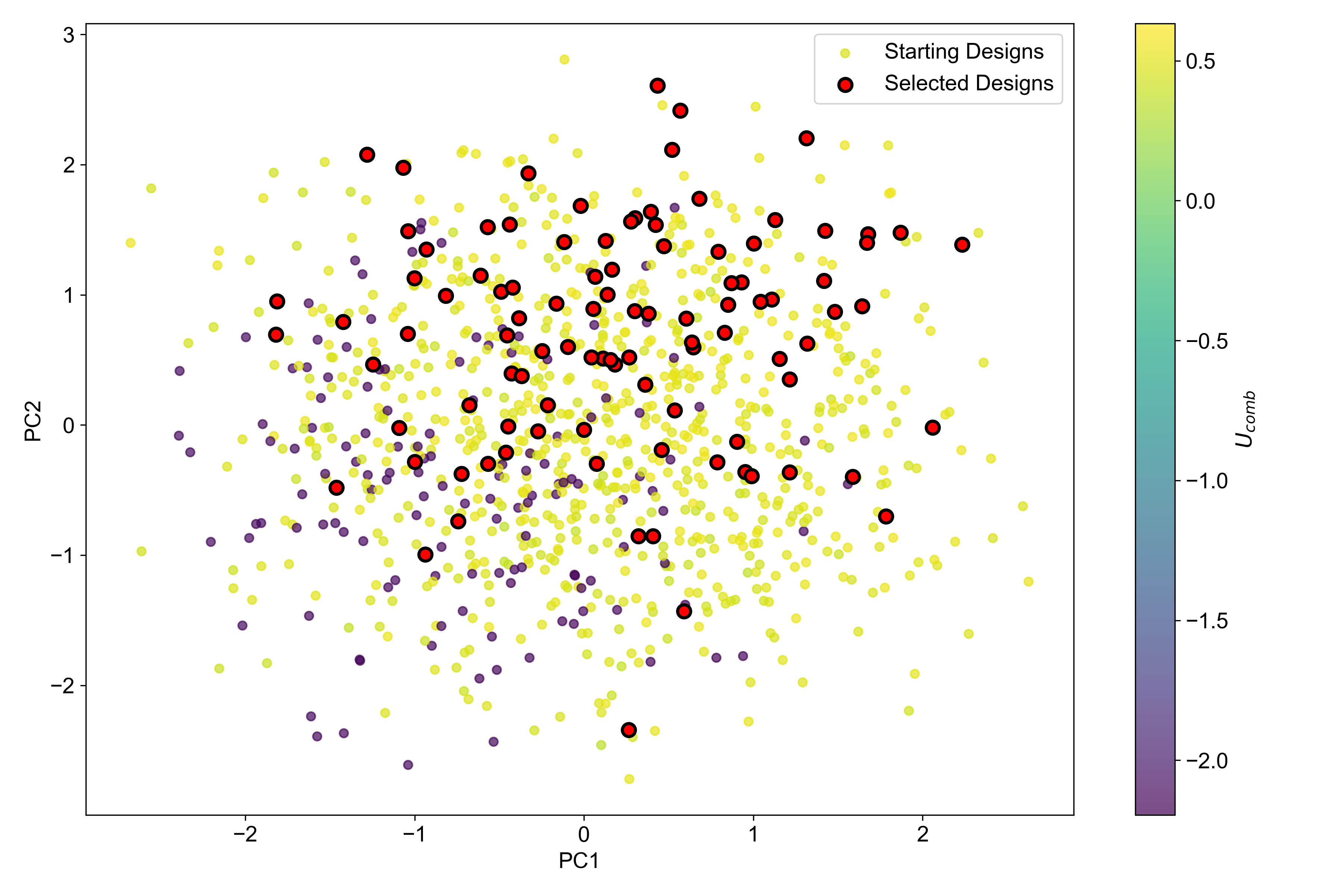}
    \caption{Plot showing the starting design set and the selected designs (in red) after the utility score-based filtering process. X and Y axes show the Principal Components of the CST parameters that define airfoil geometry. Principal Component values are used here for visualization purpose.}
    \label{fig:subset_rev0}
\end{figure}

The second stage of the workflow initiates the automated design generation and evaluation cycle, orchestrated by the Design agent. The primary objective of this phase is to efficiently explore the vast design space and identify an initial set of high-potential airfoil candidates. The process commences with the design agent systematically sampling a large set of design vectors from the predefined parameter space. Each sampled parameter set is then processed by the `AirFoil Generator' tool, which translates the abstract CST coefficients into the Cartesian coordinates defining a unique airfoil geometry. NeuralFoil surrogate is then used for rapidly predicting estimates for the key aerodynamic performance metrics: the coefficients of lift ($C_L$), drag ($C_D$), and moment ($C_M$), for each generated airfoil. Subsequently, the vector of predicted coefficients for each design is fed into the `Utility Score' tool, which converts the multi-objective performance data into a single, scalar utility score ($U_{comb}$) as previously detailed in equation \ref{eqn:utility_score}. This score provides a quantitative measure of the design's overall desirability. The final step in this stage is a critical down-selection process: the system applies a filter that automatically discards any design with a utility score below a predefined threshold of 0.4. This threshold is strategically chosen as it corresponds to the utility score of the benchmark RAE2822 airfoil, ensuring that only designs predicted to outperform the baseline are retained. In addition to this filter, we further down-sample this set by selecting the top 100 designs with highest $U_{comb}$ score. This automated filtering mechanism drastically reduces the initial candidate pool to a smaller, more manageable subset of promising designs, which then proceed to the subsequent stages of the workflow for further modification. Figure \ref{fig:subset_rev0} shows the resulting subset of designs after the utility score-based filter has been applied.

\subsection{Stage 3: Sensitivity analysis}
Following the initial design generation and filtering, the workflow transitions to a crucial analytical phase orchestrated by the Analyst agent. The primary objective of this stage is to move beyond merely identifying performant designs to fundamentally understanding the relationship between the design variables and the resulting aerodynamic performance. Given that the mapping from the nine-dimensional CST parameter space $\in \mathbb{R}^9$ to the aerodynamic coefficients ($C_L, C_D, C_M$) is highly non-linear and non-intuitive, a systematic analysis is essential to enable intelligent design refinement. To achieve this, the Analyst agent is tasked with performing a global sensitivity analysis (GSA). For this purpose, a variance-based method utilizing Sobol indices was selected, as it excels at quantifying the influence of input parameters in complex models by accounting for both their individual effects and their interactions, a significant advantage over local, one-at-a-time sensitivity measures \cite{sensitivity_analysis}. The ultimate goal is to distill the quantitative results of the GSA into a set of qualitative, actionable guidelines that can be used by the Design Agent to strategically modify airfoil geometries in subsequent stages.

To provide a quantitative foundation for these guidelines, the Sobol method decomposes the total variance of a model output, $V(Y)$, into fractions attributable to each input parameter and their interactions. For a model $Y=f(X_1, ..., X_D)$, the first-order Sobol index ($S_i$) for an input $X_i$ is defined as 
\begin{equation}
    S_i = V[E(Y|X_i)] / V(Y),
\end{equation}
which represents the main effect of $X_i$ on the output variance. Higher-order interactions are also captured; for instance, the second-order index, 
\begin{equation}
    S_{ij} = (V[E(Y|X_i, X_j)] - V[E(Y|X_i)] - V[E(Y|X_j)]) / V(Y)
\end{equation}
quantifies the portion of variance due to the interaction between $X_i$ and $X_j$ alone. To capture the full influence of a parameter, including all its interactions, the total-effect index ($S_{Ti}$) is computed as 
\begin{equation}
    S_{Ti} = E[V(Y|X_{\sim i})] / V(Y)
\end{equation}

where $X_{\sim i}$ denotes the set of all input parameters except $X_i$ \cite{sensitivity_analysis}. In our implementation, the Analyst agent applies this method to the $P=9$ CST parameters for each of the three performance metrics ($C_L, C_D, C_M$). These metrics were obtained using NeuralFoil surrogate. A set of $N=128$ base samples is drawn from the design space, and following the Saltelli sampling scheme, a total of $N(2P+2) = 2,560$ analysis points are evaluated. The agent then computes the first-order and total-effect indices, generating both graphical plots (figures \ref{fig:sobol_indx}, \ref{fig:param_effect_pairwise}) and quantitative reports. Finally, the agent processes this information to synthesize a set of strategic recommendations, such as the report shown here. These quantitative guidelines provide a structured, data-driven basis for design modification, moving the agent from a purely heuristic to an analytical approach and thereby enabling more efficient and targeted design improvement.

\begin{figure}[h]
    \centering
    \begin{textbox}{Sensitivity analysis report from Analyst}
    \scriptsize
    \begin{itemize}[leftmargin=0.2in]    
        \item \textbf{CL analysis:} Based on the Sobol analysis, the following CST parameters are most significant for the Coefficient of Lift (CL)
        \begin{itemize}
            \vspace{-5pt}\item CST\_U2: This is a highly sensitive parameter. The influence plot shows a positive correlation, meaning *increasing CST\_U2 will increase CL*.
            \vspace{-5pt}\item CST\_L3: This is another highly sensitive parameter. The influence plot shows a negative correlation, meaning *increasing CST\_L3 will decrease CL*.
            \vspace{-5pt}\item CST\_L2: This parameter has high sensitivity. The influence plot indicates a negative correlation, so *increasing CST\_L2 will decrease CL*.
            \vspace{-5pt}\item CST\_U3: This parameter has a significant impact. The influence plot shows a positive correlation, meaning *increasing CST\_U3 will increase CL*
        \end{itemize}

        \item \textbf{CD analysis:} Based on the Sobol analysis, the following CST parameters are most significant for the Coefficient of Drag (CD)
        \begin{itemize}
            \vspace{-5pt}\item CST\_L3: This is the most dominant parameter affecting CD. The influence plot shows a strong positive correlation, meaning *increasing CST\_L3 will significantly increase CD*.
            \vspace{-5pt}\item CST\_U3: This parameter is highly sensitive. The influence plot shows a negative correlation, so *increasing CST\_U3 will decrease CD*.
           \vspace{-5pt} \item CST\_U4: This parameter is also highly sensitive. The influence plot indicates a negative correlation, meaning *increasing CST\_U4 will decrease CD*.
            \vspace{-5pt}\item CST\_U2: This parameter has a moderate but notable sensitivity. The influence plot shows a negative correlation, meaning *increasing CST\_U2 will decrease CD*
        \end{itemize}

        \item \textbf{CM analysis:} Based on the Sobol analysis, the following CST parameters are most significant for the Coefficient of Moment (CM)
        \begin{itemize}
            \vspace{-5pt}\item CST\_L3: This is the most sensitive parameter for CM. The influence plot shows a positive correlation, meaning *increasing CST\_L3 will increase CM* (make it less negative).
            \vspace{-5pt}\item CST\_U2: This parameter is highly sensitive. The influence plot indicates a positive correlation, so *increasing CST\_U2 will increase CM*.
            \vspace{-5pt}\item CST\_U3: This parameter has a high sensitivity. The influence plot shows a negative correlation, meaning *increasing CST\_U3 will decrease CM*.
            \vspace{-5pt}\item CST\_L2: This parameter is also highly sensitive. The influence plot shows a positive correlation, meaning *increasing CST\_L2 will increase CM*.
        \end{itemize}
    \end{itemize}
    \end{textbox}
    \label{fig:sensitivity_report}
\end{figure}

\subsection{Stage 4: Design revision and risk-based filtering}
\begin{figure}[h]
    \centering
    \includegraphics[width=1\linewidth]{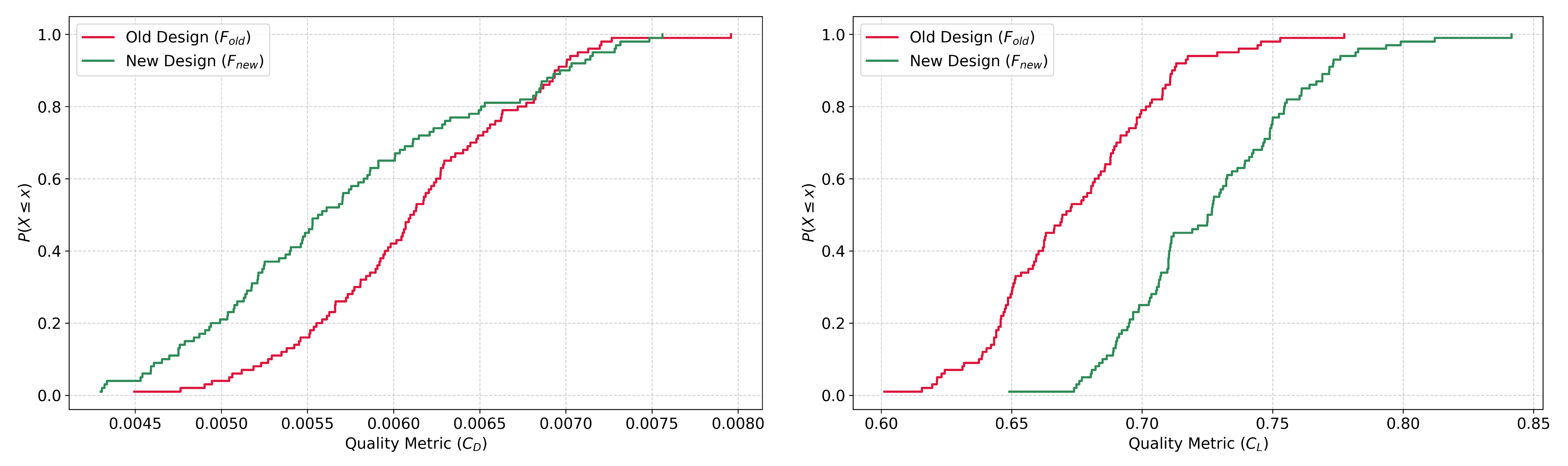}
    \caption{Comparison between cumulative distribution of $C_D$ and $C_L$ before and after design modification by the Design agent. The cumulative distribution shows the improvement in overall values of $C_D$ and $C_L$ after parameter modification based on the Sensitivity analysis.}
    \label{fig:cdf_cl_cd}
\end{figure}

Upon completion of the sensitivity analysis, the workflow advances to a directed design refinement stage, where the Design agent's objective is to enhance the performance of the 100 promising candidates selected in Stage 2. Leveraging the quantitative guidelines from the Analyst agent's sensitivity report, the Design agent systematically modifies the CST parameters of each airfoil. These modifications are targeted at the parameters identified as having the most significant influence on the aerodynamic coefficients, with the explicit goals of increasing the lift coefficient ($C_L$), reducing the drag coefficient ($C_D$), and improving the pitching moment ($C_M$). As illustrated in the comparative Cumulative Distribution Function (CDF) plots shown in the figure \ref{fig:cdf_cl_cd}, this guided refinement strategy successfully shifts the population of designs towards improved overall performance, particularly in $C_L$ and $C_D$. However, we also observe a common challenge, where improvements in lift and drag can sometimes have an adverse effect on the pitching moment. Following these modifications, the `Airfoil Generator` tool is used to create the updated airfoil geometries. At this stage, we transition from the `NeuralFoil' model to a custom-trained Bayesian surrogate for performance evaluation. This change is motivated by two key advantages: the Bayesian surrogate offers a more reliable estimate of $C_D$ within our specific design space, and more importantly, it provides a full posterior predictive distribution for each aerodynamic coefficient.

\begin{figure}[h]
    \centering
    \includegraphics[width=1\linewidth]{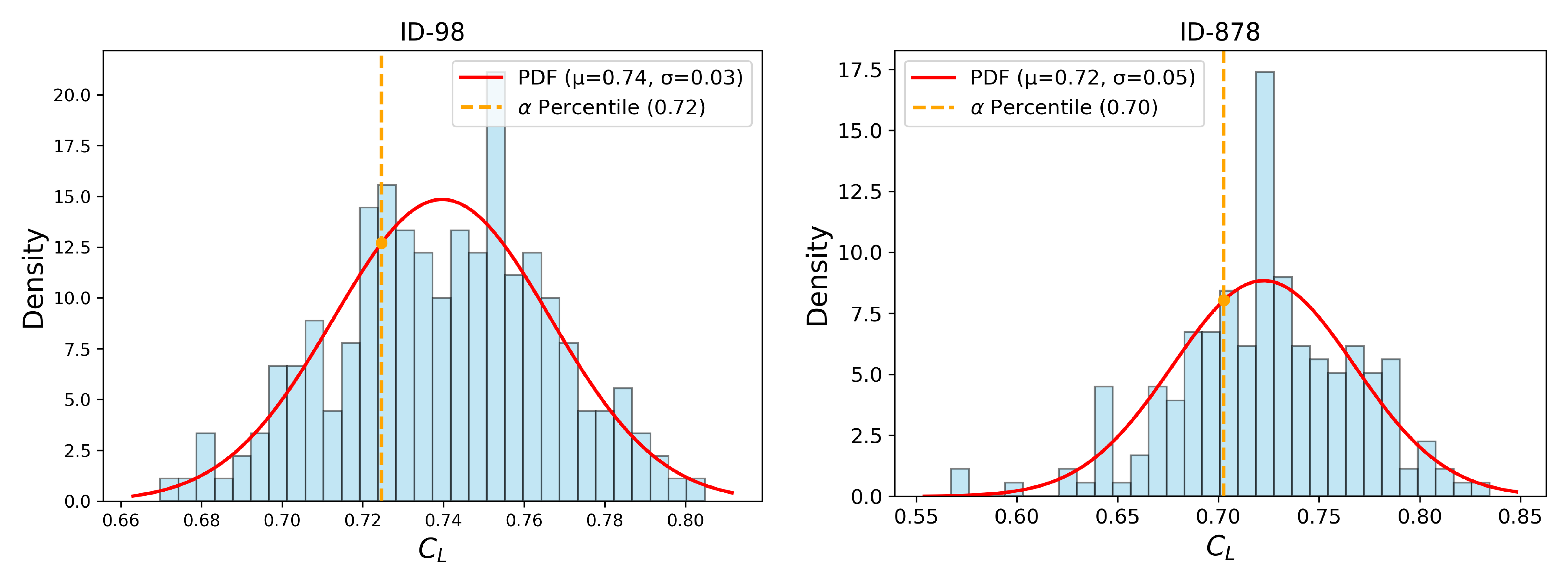}
    \caption{Plot showing $C_L$ distributions of two design candidates: ID--98 (retained) and ID--878 (removed) used for $\alpha$-risk filtering strategy. The orange line marks the $C_L$ value at $1-\alpha$ percentile of the distribution.}
    \label{fig:risk_filter_hist}
\end{figure}

The generation of refined designs is followed by a subsequent filtering step to ensure that the selected candidates are not only high-performing but also robust to the inherent uncertainties of the surrogate models. With the probabilistic performance estimates provided by the Bayesian surrogate, we now implement our risk-based filtering strategy. This approach moves beyond considering only the mean predicted performance and explicitly addresses the risk due to epistemic uncertainty in the model. Specifically, the newly generated designs are subjected to the $\alpha$-risk filter, which implements the Conditional Value-at-Risk (CVaR) methodology detailed in Algorithm \ref{alg:cvar}. For this filtering process, we define a confidence level of $\alpha=0.7$ and a target Value at Risk for the lift coefficient of $VaR_{CL}=-0.70$. This configuration ensures that for a design to be selected, its expected $C_L$ in the worst 30\% of predicted outcomes must be greater than or equal to 0.70. Figure \ref{fig:risk_filter_hist} shows the $C_L$ distribution of two design candidates; ID--98 which was selected as valid during the risk-filtering process whereas ID--878 was removed, based on CVaR value. By imposing this stringent condition, the system effectively mitigates the risk of selecting designs that appear promising due to favorable but uncertain surrogate predictions. This process prunes the set of modified designs, retaining only those that demonstrate robust performance, thereby increasing confidence in the final set of candidates passed to subsequent stages. A total of 64 designs remain at the end of Stage 4.

\subsection{Stage 5: Design Revision and evaluation of Pressure distribution}
In stge 5, the Design agent operates on the subset of designs identified as robust to epistemic uncertainty at the conclusion of Stage 4. This phase represents a shift towards fine-tuning an already promising and reliable set of candidates. The Design agent initiates a second round of modifications, once again leveraging the quantitative insights from the sensitivity analysis report generated in Stage 3. However, unlike the broader refinement in the previous stage, these modifications are more focused, with the primary objective of further enhancing the lift coefficient ($C_L$) while maintaining the favorable drag and moment characteristics of the robust designs. Following each modification, the `Airfoil Generator` tool produces the new geometry, which is then evaluated using the Bayesian surrogate. For this performance assessment, the point estimate for each aerodynamic coefficient is taken as the mean of the posterior predictive distribution, calculated across 200 independent and identically distributed (iid) ensembles. This is followed by a final round of utility scoring and filtering, mirroring the process from Stage 2. The refined designs are ranked in descending order based on their new utility scores, and the top 50 designs are retained, forming the subset for detailed review.

\begin{figure}[h]
    \centering
    \includegraphics[width=0.8\linewidth]{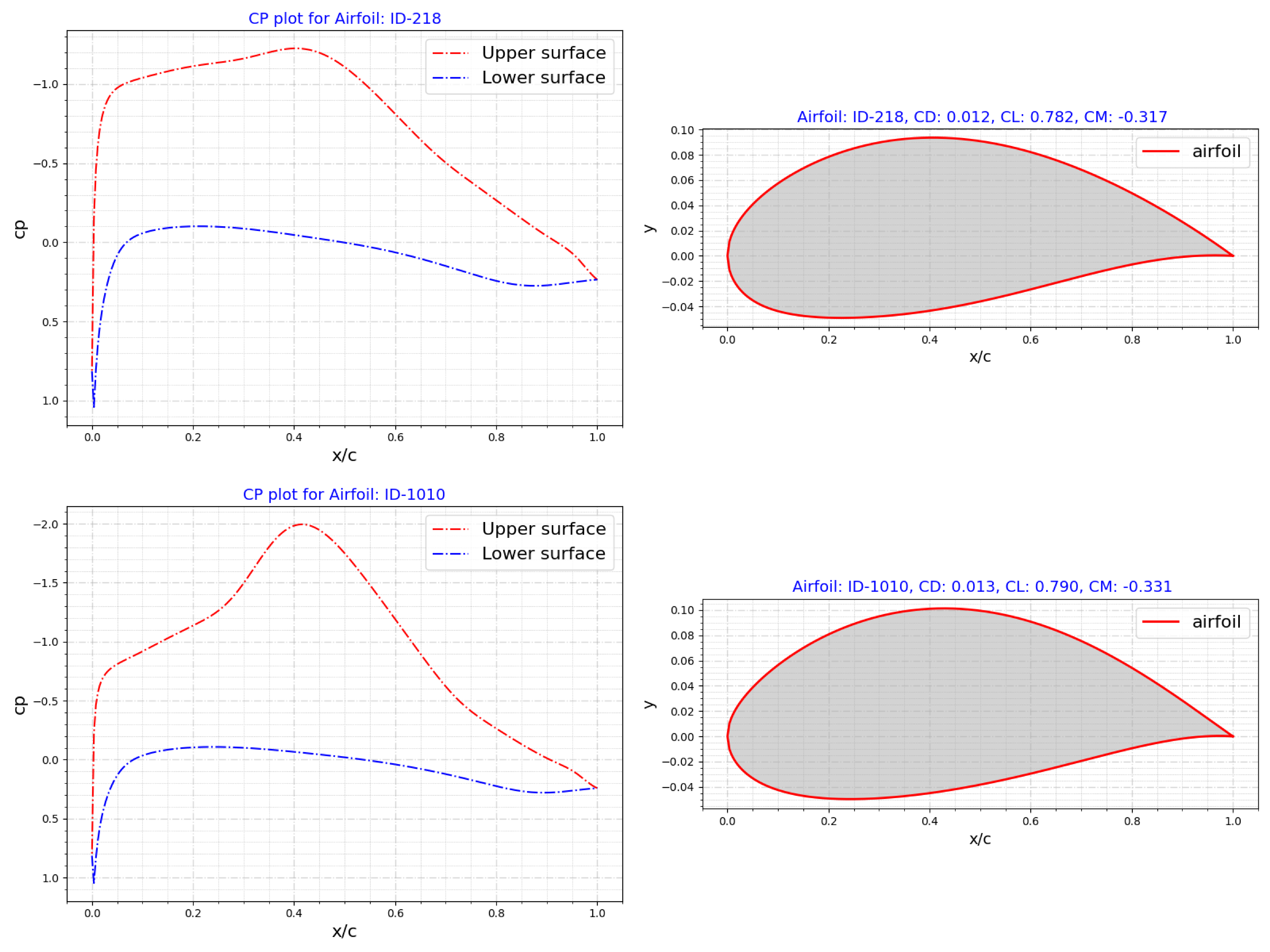}
    \caption{Coefficient of Pressure plots for two airfoil samples generated after analysis using the DeepONet surrogate. These pressure plots allows the Systems Engineer agent to perform a comprehensive review of the airfoil design, to include choice of design parameters, integral coefficients such as $C_D, C_L$, and $C_M$, and the pressure distribution curves generated here.}
    \label{fig:cp_plots_stage4}
\end{figure}

With the final set of 50 robust designs established, the analysis transitions from evaluating integral aerodynamic coefficients to a more detailed characterization of the flow physics. This is crucial because integral values like $C_L$ and $C_D$ do not fully capture the nuanced aerodynamic behavior of an airfoil. To facilitate this, the framework employs a specialized Deep Operator Network (DeepONet) surrogate, discussed earlier in section \ref{sec:pressure_eval_tool}. The Design agent utilizes this tool to generate a $C_P$ plot for each of the 50 selected designs. A sample of this plot generated for all 50 design candidates is shown in figure \ref{fig:cp_plots_stage4}. The analysis of these distributions is important for a comprehensive design assessment, as it allows for the qualitative evaluation of critical flow features. This includes identifying the location and strength of shock waves, assessing the adversity of the pressure gradient on the aft portion of the airfoil which can indicate a propensity for flow separation, and ensuring a smooth pressure recovery. The generation of these detailed $C_P$ plots provides the necessary physical insights, which, along with the final performance metrics, serve as the comprehensive input for the subsequent automated design review and final selection stage.

\subsection{Stage 6: Automated design review and filtering}
Following the generation and detailed characterization of the top 50 designs in the preceding stage, the workflow transitions to a rigorous vetting process. This phase is executed by a specialized `Systems Engineer' agent, whose primary function is to perform a holistic evaluation that emulates the judgment of an experienced human engineer. The agent's task is to holistically assess the designs based on a combination of quantitative metrics and qualitative aerodynamic features. This is achieved through a two-pronged evaluation strategy: first, by re-evaluating the utility scores for the aerodynamic coefficients ($C_D$, $C_L$, $C_M$), and second, by assigning a qualitative rating to the airfoil's pressure distribution ($C_P$) curve. The overarching goal of this stage is to automate the final down-selection by systematically applying a set of heuristic rules to classify each of the 50 candidates as either 'valid' or 'invalid', thereby filtering out designs that are numerically optimal but aerodynamically or structurally impractical.

The evaluation rubric for the Systems Engineer agent is structured around a direct comparison with the RAE2822 benchmark airfoil (shown in Appendix figure \ref{fig:RAE2822}), whose performance at the specified flow conditions is characterized by $C_D = 0.010$, $C_L=0.522$, and $C_M=-0.073$, corresponding to utility scores of $U(C_D) = 0.518$, $U(C_L) = 0.177$, $U(C_M) = 0.7566$, and a combined score of $U_{comb} = 0.3955$. The qualitative assessment of the $C_P$ curve is performed using an ordinal rating on a scale of 1 (worst) to 5 (best), with the benchmark's pressure distribution assigned a baseline rating of 3. The agent is instructed to penalize designs exhibiting aerodynamically poor features; for instance, a significant peak or bump on the upper surface pressure distribution between 30\% and 60\% of the chord: a typical indicator of a strong, undesirable shock wave results in a low rating of 1 or 2. This automated review also includes a check against non-functional requirements such as manufacturability and structural integrity by comparing the candidate's geometry against the standard shape of the RAE2822. This multi-faceted analysis culminates in a single binary classification, governed by a strict rule: a design is deemed `Valid` only if its combined utility score is greater than or equal to 0.41 and its pressure distribution rating is 3 or higher. The specific instructions provided to the agent are summarized in figure \ref{fig:sys_engg_rules}. 

\begin{figure}[h]
    \centering
    \begin{textbox}{Design evaluation rules for Systems Engineer agent}
    \scriptsize
    \begin{enumerate}
        \vspace{-5pt}\item \textbf{Pressure Distribution ($C_P$) Rating (Scale 1-5):} Compare the sample's $C_P$ plot to the benchmark. The benchmark is rated 3. A rating of 3 or higher is assigned if the sample shows no significant adverse departure. A pronounced peak on the upper surface near the mid-chord ($0.3 < x/c < 0.6$) must be rated 1 or 2 based on severity.
        \vspace{-5pt}\item \textbf{Non-Functional Requirements:} Assess the airfoil shape for practical viability (e.g., manufacturability, structural integrity) using the benchmark shape as a reference.
        \vspace{-5pt}\item \textbf{Performance Coefficient Comparison:} Quantitatively compare the sample's $C_L$, $C_D$, and $C_M$ values against the benchmark ($C_D = 0.01$, $C_L=0.522$, $C_M=-0.073$).
        \vspace{-5pt}\item \textbf{Utility Score Comparison:} Compare the sample's combined utility score ($U_{comb}$) against the benchmark's score of 0.3955.
        \vspace{-5pt}\item \textbf{Final Validity Decision:} A design is classified as \textbf{Valid} if and only if:
            \begin{itemize}
                \vspace{-5pt}\item $U_{comb} \geq 0.41$
                \vspace{-5pt}\item $C_P$ Rating $\geq 3$
            \end{itemize}
    \end{enumerate}
    \end{textbox}
    \caption{Instruction set provided to the Systems Engineer during the automated design evaluation and filtering phase.}
    \label{fig:sys_engg_rules}
    \end{figure}

\begin{figure}[H]
    \centering
    \begin{response}{Design evaluation for Design ID-762 (Invalid)}
    \scriptsize
        ``The pressure distribution for ID-762 shows a very strong suction peak on the upper surface around x/c = 0.4, reaching approximately -2.5 Cp. This peak is significantly more pronounced and located further aft compared to the benchmark RAE2822. The lower surface pressure is relatively flat and slightly negative over a large portion, which differs from the more positive pressure observed on the benchmark's lower surface. This strong mid-chord suction peak is a concern for potential shock formation and increased drag at higher speeds. The airfoil ID-762 is noticeably thicker and more highly cambered than the RAE2822 benchmark. This increased thickness generally contributes to better structural integrity. The smooth contour suggests good manufacturability and maintainability, although the higher camber might require specific manufacturing considerations. The airfoil ID-762 achieves a significantly higher coefficient of lift (CL = 0.865) compared to the benchmark (CL = 0.522). However, it also exhibits higher drag (CD = 0.013 vs 0.010) and a substantially larger nose-down pitching moment (CM = -0.362 vs -0.073), which could pose stability challenges. The combined utility score for ID-762 is 0.4898, which is higher than the benchmark's 0.3955. This is primarily driven by a very high utility for lift (util(CL) = 0.7217). However, the utility for pitching moment (util(CM) = 0.0) is very low, indicating a significant penalty for the large negative moment, and the utility for drag (util(CD) = 0.4297) is also lower than the benchmark.''

        Pressure rating: 2; Valid: False
        
        \centering
        \includegraphics[width=0.6\linewidth]{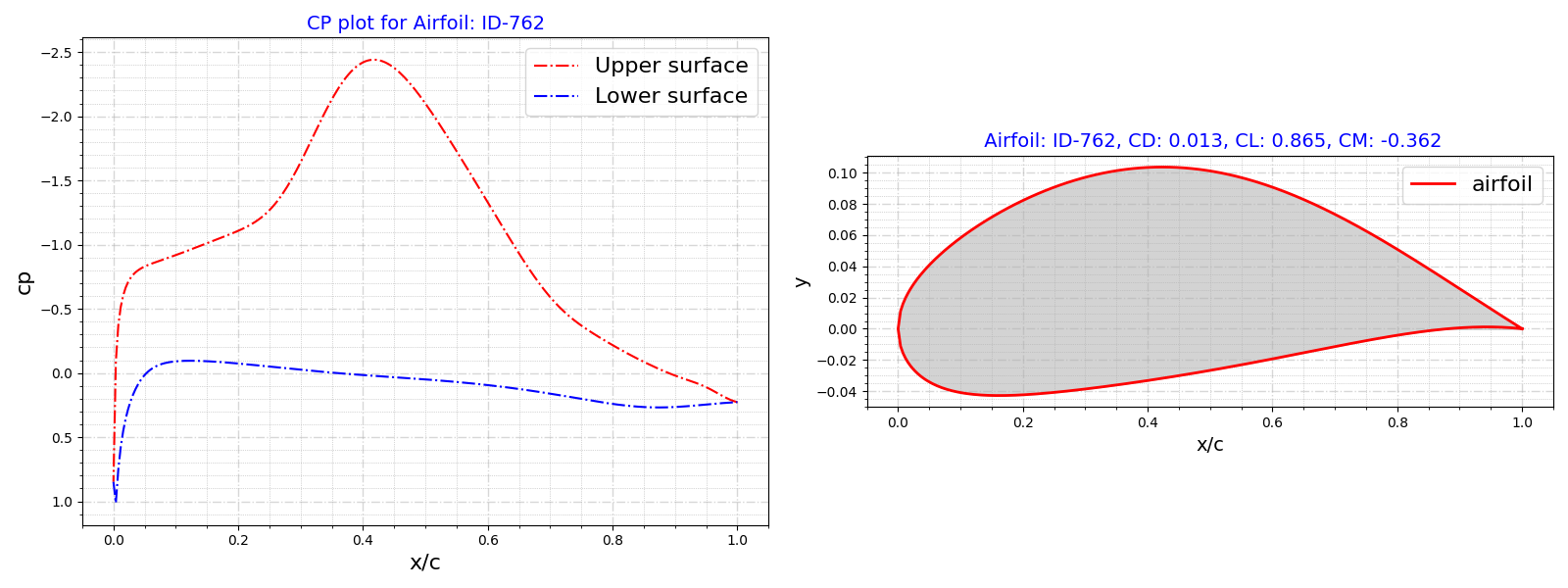}
    \end{response}
    \caption{Example of feedback generated by the Systems Engineer during automated design evaluation stage for an invalid design.}
    \label{fig:invalid_design_feedback}
    \end{figure}

\begin{figure}[h]
    \centering
    \begin{response}{Design evaluation for Design ID-470 (Valid)}
    \scriptsize
        ``The pressure distribution for airfoil ID-470 exhibits a smooth and well-behaved curve on both upper and lower surfaces, with no significant adverse pressure gradients or sudden changes, indicating good flow characteristics. The suction peak on the upper surface is less pronounced than the benchmark RAE2822, which contributes to its higher lift. The airfoil profile is robust with smooth contours and a relatively thick section, suggesting good manufacturability, structural integrity, and no obvious maintenance issues. Aerodynamically, ID-470 generates significantly more lift (CL = 0.737) than the benchmark (CL = 0.522), leading to a much higher lift utility. However, it also has a slightly higher drag (CD = 0.011 vs 0.010) and a considerably larger nose-down pitching moment (CM = -0.285 vs -0.073), which severely reduces its moment utility. Despite the low moment utility, the high lift performance results in a combined utility score (0.452) that surpasses the benchmark (0.396).''

        Pressure rating: 4; Valid: True
        
        \centering
        \includegraphics[width=0.6\linewidth]{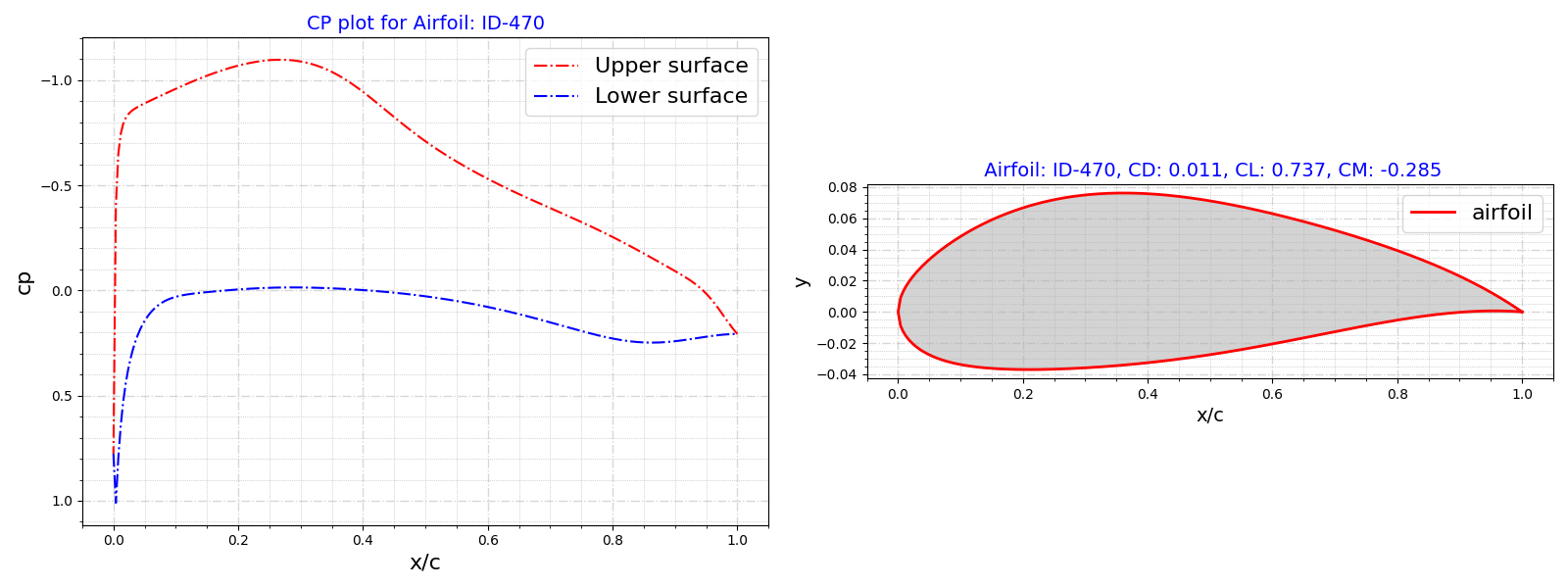}
    \end{response}
    \caption{Example of feedback generated by the Systems Engineer during automated design evaluation stage for a valid design.}
    \label{fig:valid_design_feedback}
    \end{figure}

In practice, the Systems Engineer agent systematically processes each of the 50 final design candidates, applying the defined rubric to generate the utility scores and ordinal $C_P$ ratings. This automated procedure ensures a consistent and objective application of the evaluation criteria across the entire design cohort. Subsequently, the agent applies the final filtering rule, partitioning the set of 50 designs into two distinct groups: a smaller subset of 'valid' designs that satisfy both the performance and aerodynamic quality thresholds, and a group of 'invalid' designs that are discarded in next steps. This automated rating and filtering step is a critical component of the multi-agent framework, as it codifies expert-level judgment into a repeatable process, thereby saving significant man-hours that would otherwise be spent on manual review. It is important to note that the agent's role in this final stage is purely evaluative. It does not generate feedback for design improvement; rather, its purpose is to deliver a refined, high-quality subset of designs for final consideration by the human design team. Figures \ref{fig:invalid_design_feedback} and \ref{fig:valid_design_feedback} show examples of responses generated by the Systems Engineer agent during this automated review process, to filter out designs.

\subsection{Iterative Human-in-the-Loop Design Review and Refinement}
After the automated design review and filtering stage, the workflow enters a final, collaborative phase of iterative design revision and review. This stage is initiated with the small subset of designs classified as 'Valid' by the Systems Engineer agent in the preceding step. The core objective shifts from broad exploration and automated filtering to a highly focused, human-supervised refinement process. This human-in-the-loop approach is critical for preventing 'runaway' design choices, where an autonomous system might over-optimize for a specific metric at the expense of unquantified but crucial engineering characteristics. The process involves a tight feedback loop between the Design agent, the Systems Engineer agent, and a human Manager, ensuring that the final designs are not only performant according to the defined user goals but also align with the nuanced, often tacit, requirements of a practical engineering application.

\begin{figure}[h]
    \centering
    \begin{response}{Design feedback for Design ID-486 - iteration 1}
    \scriptsize
        ``The pressure distribution for airfoil ID-486 shows a generally similar trend to the RAE2822 benchmark, with an initial acceleration to low pressure on the upper surface. However, there is a slight irregularity or less smooth pressure recovery on the upper surface between x/c = 0.2 and x/c = 0.4, which is a minor departure from the benchmark's smoother curve. The lower surface pressure distribution is comparable to the benchmark, with a slightly more pronounced pressure minimum near the leading edge. Overall, the pressure distribution is acceptable. The airfoil shape of ID-486 appears robust and well-proportioned, similar in general characteristics to the RAE2822 benchmark. Visually, it does not present any features that would raise concerns regarding manufacturability, structural integrity, or maintenance. The thickness distribution and camber seem appropriate for an efficient airfoil design. 
        
        Comparing aerodynamic performance, the coefficient of drag (CD) for ID-486 is slightly higher than the benchmark (0.01038 vs 0.010), resulting in a slightly lower utility score for drag (0.509 vs 0.518). The coefficient of lift (CL) is significantly higher for ID-486 (0.735 vs 0.522), leading to a much better utility score for lift (0.579 vs 0.177). The coefficient of moment (CM) for ID-486 is considerably more negative (-0.289 vs -0.073), which severely impacts its utility score for moment (0.0375 vs 0.7566), indicating a strong nose-down pitching moment. Despite the poor CM utility, the high CL utility helps the combined utility score of ID-486 (0.450) to be higher than the benchmark (0.3955).''

        Pressure rating: 3; Valid: True
        
        \centering
        \includegraphics[width=0.6\linewidth]{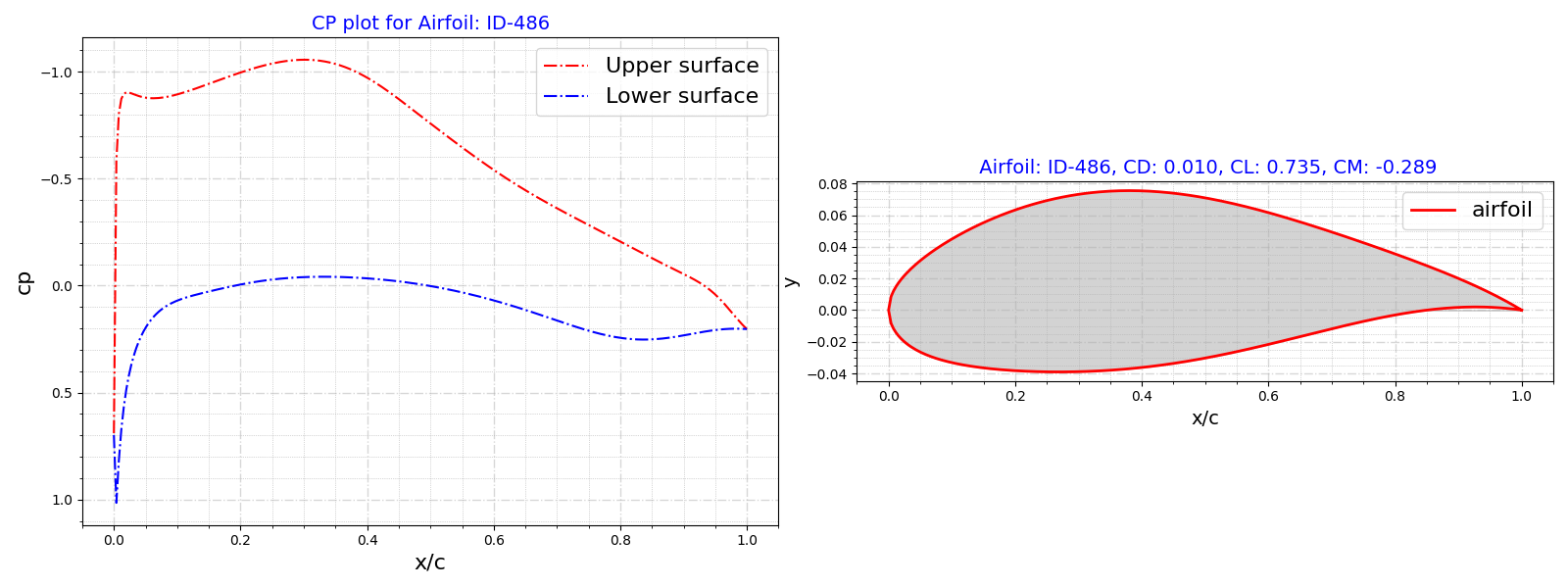}
    \end{response}
    \caption{Feedback for design ID-486 generated by the Systems Engineer with human Manager during first iteration.}
    \label{fig:design_feedback_iter1}
\end{figure}

\begin{figure}[h]
    \centering
    \begin{response}{Design feedback for Design ID-486 - iteration 2}
    \scriptsize
        ``\textbf{Pressure Distribution}: The airfoil ID-486 exhibits a strong suction peak on the upper surface near the leading edge, reaching Cp values below -1.0, which contributes to high lift. However, there is a noticeable 'bump' around x/c = 0.2, indicating potential local flow acceleration or a weak shock, and the pressure recovery on the upper surface is less smooth compared to the benchmark RAE2822. The lower surface pressure distribution is generally smooth and positive.
        Airfoil Profile: Airfoil ID-486 appears thicker and has more pronounced camber than the benchmark RAE2822, particularly in the mid-chord region. The leading edge radius seems larger. The overall shape appears structurally sound and manufacturable, similar to the benchmark, without any extreme features that would raise immediate concerns regarding manufacturability, structural integrity, or maintenance.
        
        \textbf{Aerodynamic Performance \& Utility}: The airfoil ID-486 achieves a significantly higher Coefficient of Lift (CL = 0.735) compared to the benchmark (CL = 0.522), resulting in a much higher CL utility score (0.5800 vs 0.177). The Coefficient of Drag (CD = 0.010) is comparable to the benchmark (CD = 0.010), leading to a similar CD utility score (0.5104 vs 0.518). However, the Coefficient of Moment (CM = -0.287) is substantially more negative than the benchmark (CM = -0.073), resulting in a very low CM utility score (0.0425 vs 0.7566). Despite the poor CM utility, the high CL utility leads to a higher combined utility score (0.4516) for ID-486 compared to the benchmark (0.3955).
        
        \textbf{Design Improvements:}
        \begin{enumerate}
            \vspace{-5pt}\item Increase CST\_U2: This is the most impactful change. Increasing `CST\_U2' will simultaneously increase CM (making it less negative), increase CL, and decrease CD, thereby improving the overall combined utility and pitching moment characteristics.
            \vspace{-5pt}\item Carefully decrease CST\_U3: To further improve CM (make it less negative), a controlled decrease in `CST\_U3' can be explored. However, this must be balanced against its negative impact on CL and positive impact on CD.
            \vspace{-5pt}\item Consider increasing CST\_L2: A moderate increase in `CST\_L2' could help increase CM. This would come at the cost of decreasing CL, so it should be considered if the current high CL can be slightly reduced for a more favorable pitching moment.
        \end{enumerate}
        Pressure rating: 3; Valid: True ''
        
        \centering
        \includegraphics[width=0.6\linewidth]{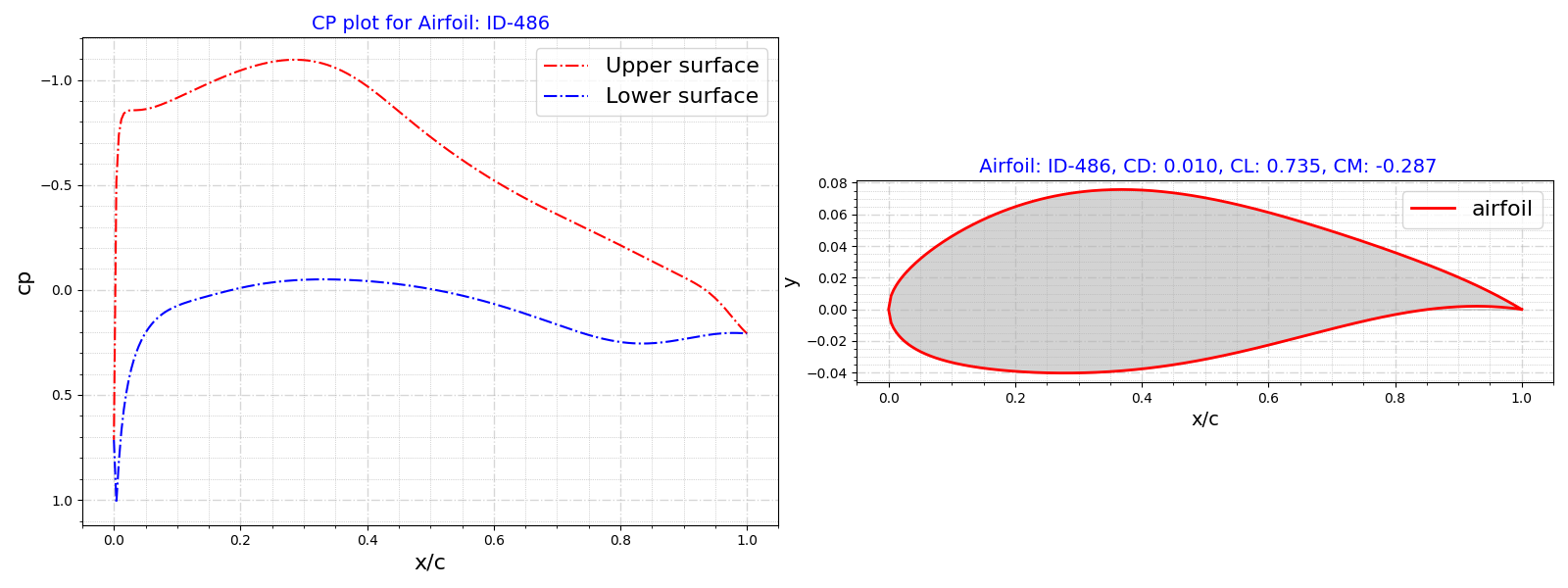}
    \end{response}
    \caption{Feedback for design ID-486 generated by the Systems Engineer with human Manager during second iteration after design modification.}
    \label{fig:design_feedback_iter2}
\end{figure}

Each iteration within this phase begins with the Design agent modifying a candidate airfoil. The agent's modifications are informed by a combination of two key inputs: the global sensitivity analysis report, which provides a foundational understanding of the parameter-performance relationship, and specific, targeted feedback generated from the previous review cycle. Once a design is modified, its new geometry and predicted pressure distribution are presented for a joint review by the Systems Engineer agent and the human Manager. The Systems Engineer agent first performs an assessment analogous to its role in Stage 6, recalculating the combined utility score and assigning a qualitative rating to the pressure curve. This provides a consistent, data-driven baseline for the review. However, the final authority in this stage rests with the human Manager, who can override the agent's assessment. If the Manager approves a design as a candidate for further improvement, the Systems Engineer agent is tasked with generating specific, actionable feedback. This feedback leverages both the sensitivity data and the agent's analysis of the current design's shortcomings to guide the Design agent's next modification (e.g., "Reduce the adverse pressure gradient post-shock by adjusting the aft-camber parameters"). Figures \ref{fig:design_feedback_iter1} -- \ref{fig:design_feedback_iter2} show the feedback generated by the Systems Engineer for the first two design loops for a particular design that remains `Valid' till the end of the design cycle. Additional responses for the third and fourth iteration can be found in figures \ref{fig:design_feedback_iter3} and \ref{fig:design_feedback_iter4} in Appendix. Note that after every feedback, the Design agent modifies the design to improve the particular aspect of design where design improvements were suggested. In general, all the designs selected till the end exhibit high $C_L$ values but suffer from negative $C_M$ values, hence the design improvement strategy mostly focuses on improving $C_M$ in the iterative stage. 

\begin{figure}[h]
    \centering
    \includegraphics[width=1\linewidth]{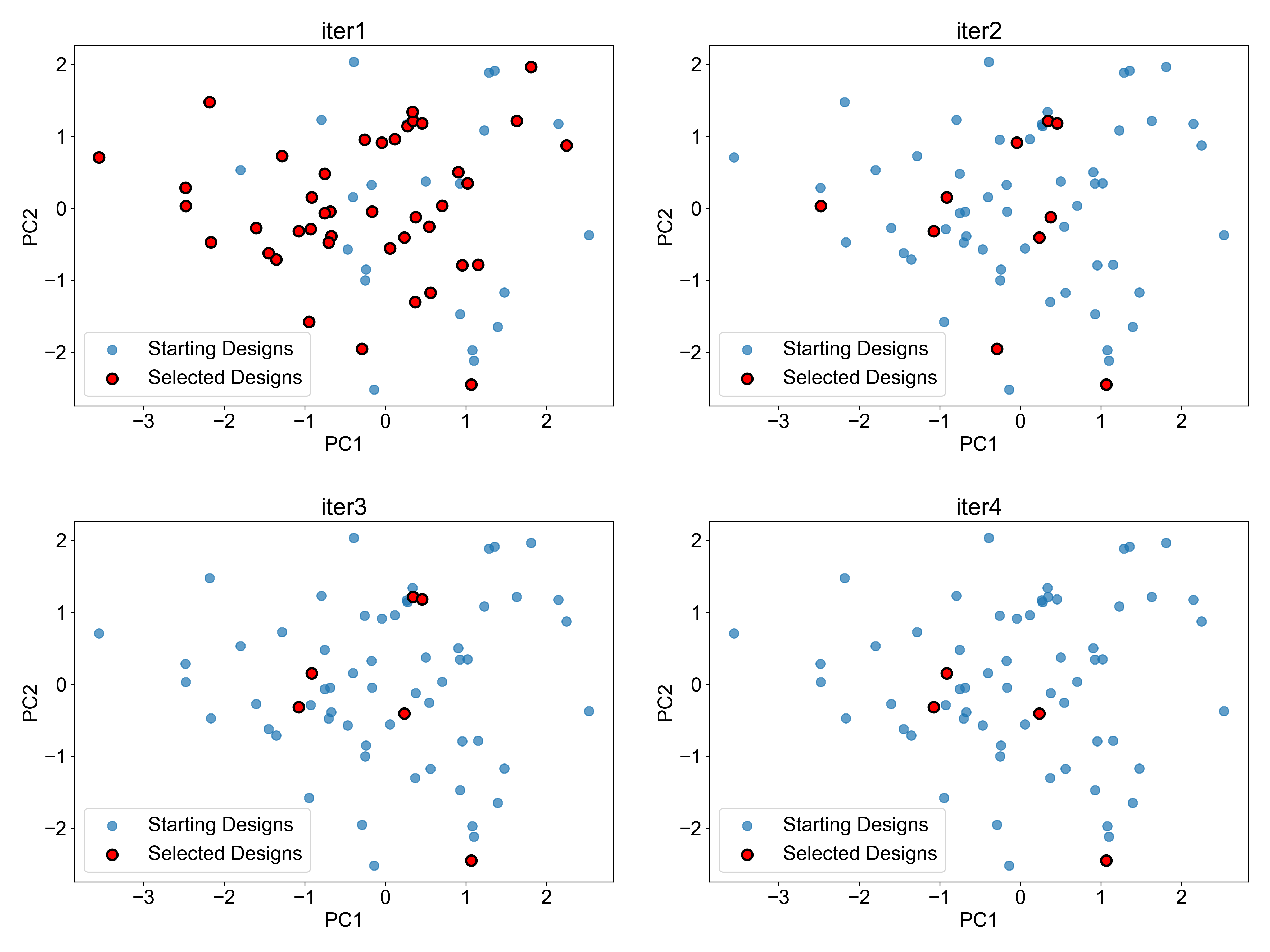}
    \caption{Plot showing the design filtering during iterative design and review process. The x and y-axes show the two principal components for the CST parameters for each design. In the iterative phase, the designs are filtered based on the human Manager's input along with feedback received from the Systems Engineer agent.}
    \label{fig:design_filtering}
\end{figure}

This iterative cycle of modification, joint review, and feedback generation continues until a final, small subset of approximately four--five designs is identified, meeting the satisfaction of the human Manager. The successive filtering of designs is shown in figure \ref{fig:design_filtering}. The determination of design validity in this phase is explicitly deferred to human judgment, given the significantly reduced number of candidates, which makes manual review non-labor-intensive. Once this set of designs is finalized, the Manager has the option to trigger a final verification step to confirm the suitability of the candidates with a numerical CFD model. This serves as a crucial final validation, ensuring that the performance predicted by the surrogate models translates accurately into a physics-based simulation environment before final design selection.

\subsection{Stage 7: CFD simulation with OpenFoam}
The iterative design cycle culminates when the Manager identifies a small, curated subset of designs deemed suitable for the application. At this juncture, the workflow transitions from rapid, surrogate-based evaluation to high-fidelity physical verification. The Manager formally requests a Computational Fluid Dynamics (CFD) evaluation for each of the final candidate airfoils. In response, the Systems Engineer agent initiates a series of automated simulation tasks using the open-source OpenFOAM framework. The core of this analysis is the `simpleFoam' solver, a steady-state solver for incompressible and compressible flows. The simulations solve the Reynolds-Averaged Navier-Stokes (RANS) equations to model the fluid dynamics. Specifically, for a compressible fluid treated as a perfect gas, the solver iteratively finds a solution to the steady-state continuity and momentum equations:

\begin{align}
    &\nabla \cdot (\rho \mathbf{U}) = 0 \\
    &\nabla \cdot (\rho \mathbf{U} \mathbf{U}) = -\nabla p + \nabla \cdot \boldsymbol{\tau}_{eff}
\end{align}

Here, $\rho$ is the fluid density, $\mathbf{U}$ is the mean velocity vector, $p$ is the mean pressure, and $\boldsymbol{\tau}_{eff}$ is the effective stress tensor, which includes both molecular and turbulent stresses. To close this system of equations, turbulence is modeled using the one-equation Spalart-Allmaras model, a choice well-suited for external aerodynamic applications involving attached boundary layers, as is common in airfoil analysis. This model introduces a transport equation for a modified turbulent kinematic viscosity, thereby providing the turbulent viscosity term required to compute the turbulent stresses within $\boldsymbol{\tau}_{eff}$.

To facilitate an automated and repeatable simulation process for various airfoil geometries, a series of scripts developed in prior work were used to handle the mesh generation pipeline. The process begins with a Python script that converts the 2D coordinates of a given airfoil design into a 3D surface geometry file in the Wavefront OBJ format, providing a small extrusion in the spanwise direction. This OBJ file serves as the geometric definition of the airfoil surface for the meshing utility. The subsequent meshing is handled by OpenFOAM's native `blockMesh' tool, which is configured via a templated `blockMeshDict' file. This dictionary defines a structured C-grid topology around the airfoil, a standard and efficient approach for resolving the key flow features in external aerodynamics. The mesh is parameterized to allow for programmatic updates; another Python script automatically modifies the `blockMeshDict' to adjust vertex locations corresponding to the leading edge, trailing edge, and points of maximum thickness for each new airfoil design. Significant mesh grading is applied, particularly in the direction normal to the airfoil surface, to achieve a high density of cells within the boundary layer, which is critical for accurately predicting skin friction drag and flow separation characteristics.

\begin{figure}[h]
    \centering
    \includegraphics[width=1\linewidth]{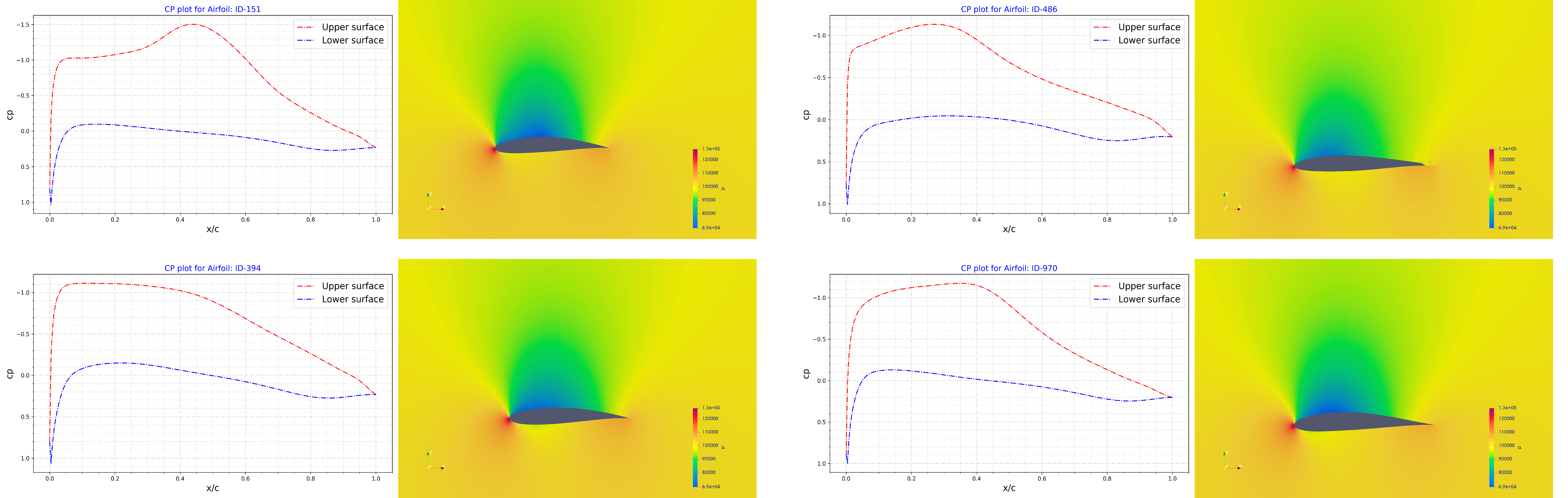}
    \caption{Coefficient of pressure plots predicted by the agentic framework and the corresponding pressure distribution plot from OpenFoam simulation for the final four design candidates.}
    \label{fig:cp_cfd_comparison}
\end{figure}

Upon completion of the CFD simulations for all selected candidates, the high-fidelity results are presented to the Manager for final review and down-selection. Figure \ref{fig:cp_cfd_comparison} shows a comparison between coefficient of pressure distribution predicted by the agentic framework and the corresponding CFD pressure distribution for the final four design candidates. Additional velocity field plots for these candidates can be found in Appendix figure \ref{fig:cfd_velocity}, which are used for qualitative analysis by the Manager. This final assessment focuses on comparing the pressure distributions and integral aerodynamic coefficients ($C_L$, $C_D$, $C_M$) against the predictions from the surrogate models used in the earlier stages. This step serves two critical functions: first, it provides validation of a design's performance in a physics-based environment, and second, it verifies the accuracy of the surrogate models used throughout the design process. For instance, in one review, design ID-151 was discarded due to a strong suction peak observed near the mid-chord in its CFD-computed pressure plot, a feature that was hinted at by the surrogate's $C_P$ prediction but was confirmed to be unacceptable by CFD analysis. Similarly, ID-486 was deemed as invalid due to its low $C_L$ value, although this design has the best $C_M$ value amongst this lot. The remaining two design candidates exhibited favorable aerodynamic performance and pressure distributions, aligning well with the surrogate predictions, and were thus retained for further development. The reasonable agreement between surrogate and CFD results, as summarized in the comparison table \ref{tab:coeff_surrogate_vs_CFD}, validates the efficacy of the multi-agent workflow in identifying high-quality designs prior to expensive, physics-based simulations.

\begin{table}[h]
\centering
\caption{Comparison between aerodynamic performance coefficient prediction using surrogate models at final design iteration against CFD predicted coefficients. We note that the surrogate model provides reasonable estimation of aerodynamic coefficients, $C_D$ and $C_L$ for these four design candidates. However, the epistemic uncertainty ($\sigma$) when evaluating these final designs is high, as compared to the original training dataset. As a consequence of design modifications performed on the airfoil shapes, the CST parameters of the final candidates reside on the periphery of training data distribution leading to higher epistemic uncertainty. This bias is mitigated, to a certain extent, in the design selection strategy by choosing a composite utility score.}
\label{tab:coeff_surrogate_vs_CFD}
\resizebox{0.9\linewidth}{!}{%
\begin{tabular}{l!{\vrule width \lightrulewidth}ccc!{\vrule width \lightrulewidth}ccc} 
\toprule
\multicolumn{1}{c!{\vrule width \lightrulewidth}}{\textbf{Design ID}} & \multicolumn{3}{c!{\vrule width \lightrulewidth}}{\textbf{Surrogate}} & \multicolumn{3}{c}{\textbf{CFD}} \\ 
\midrule
 & \multicolumn{1}{c}{$C_D \; (\sigma = 0.0011)$} & \multicolumn{1}{c}{$C_L \;(\sigma=0.024)$} & \multicolumn{1}{c!{\vrule width \lightrulewidth}}{$C_M \; (\sigma = 0.012)$} & \multicolumn{1}{c}{$C_D$} & \multicolumn{1}{c}{$C_L$} & \multicolumn{1}{c}{$C_M$} \\
 \midrule
ID-151 (invalid) & 0.013 & 0.836 & -0.348 & 0.016 & 0.88 & -0.145 \\
ID-394 (valid) & 0.012 & 0.761 & -0.307 & 0.0159 & 0.813 & -0.128 \\
ID-486 (invalid) & 0.01 & 0.731 & -0.283 & 0.02 & 0.6682 & -0.087 \\
ID-970 (valid) & 0.011 & 0.741 & -0.279 & 0.018 & 0.713 & -0.097 \\
\bottomrule
\end{tabular}
}
\end{table}

Eventually, two design candidates: ID-394 and ID-970 are identified as potential solutions for this problem. Hereafter, these candidates can be used for detailed design that may include more advanced CFD simulation prior to design optimization and detailed design phase. We consider these steps as beyond the scope of our current work.

\section{Limitations}
While this work demonstrates a successful application of a multi-agent framework to a canonical engineering design problem, we acknowledge some limitations that offer avenues for future research. Firstly, the overall design workflow is orchestrated and supervised by a human manager. The agents function effectively as specialized assistants within a predefined structure, executing tasks such as design modification, sensitivity analysis, and performance vetting, but the high-level process logic and strategic decisions remain within the human domain. Consequently, this study does not demonstrate a fully autonomous, end-to-end design process, which would require agents capable of dynamic workflow planning and adaptation. A second limitation concerns the agents' ability to develop tools in-situ. We observed that the autonomous generation and integration of new analysis or utility scripts during the workflow was a challenge, often requiring numerous rounds of human-guided iteration to debug and rectify. The framework proved far more robust when supplied with a pre-vetted, comprehensive suite of tools prior to execution, indicating that while current LLM-based agents excel at tool use, reliable on-the-fly tool creation in complex engineering contexts remains an open research opportunity. Finally, the scope of the CFD validation is constrained by the geometric complexity of the test case. Our automated CFD pipeline, particularly the meshing process which utilizes a structured C-grid topology, is tailored for 2D airfoil sections. Its direct applicability to more complex 3D geometries, such as wing-body junctions or internal turbine passages, which often necessitate advanced unstructured or hybrid meshing techniques, has not been explored. Some recent work such as \cite{openfoamgpt} have explored using multi-agent setup for automating CFD flows, and could provide a future pathway to comprehensive design automation.

\section{Summary}
In this work, we build upon our prior research to introduce a multi-agent framework, guided by Large Language Models (LLMs), for assisting in the design and evaluation of engineering systems. We demonstrate this framework on the canonical problem of airfoil design, integrating it within a set-based design philosophy that incorporates formal risk management principles. The methodology employs a workflow managed by a human expert and executed by four specialized agents: a Coding Assistant, a Design Agent, a Systems Engineering Agent, and an Analyst Agent, each tasked with a specific function. Initially, the human Manager collaborates with the Coding Assistant to define the operational workflow and develop a suite of validated computational tools. Following this setup, the agents engage in a systematic process to progressively narrow a large set of potential design candidates, using the pre-developed tools and agent-based assessments to prune the design space.

A key contribution of our approach is the explicit integration of risk assessment into the automated design process. We adopt the Conditional Value-at-Risk (CVaR) as a risk metric to quantitatively filter out design candidates that exhibit a high probability of failing to meet the required design goal for the coefficient of lift ($C_L$). The framework automates the initial, labor-intensive stages of design exploration by having the Analyst agent generate a global sensitivity analysis report. This report provides actionable heuristics that guide the Design and Systems Engineering agents in the simultaneous analysis and modification of multiple designs. The process culminates in a human-in-the-loop stage where the human Manager acts as the final decision-maker. The framework augments the Manager's capability by presenting a curated final set of design candidates, along with high-fidelity Computational Fluid Dynamics (CFD) simulation results, to inform the ultimate selection. Looking forward, future work will focus on two primary directions: first, enhancing agent autonomy to enable dynamic workflow adaptation in response to real-time findings, and second, extending the framework's application to more complex 3D multi-physics problems, which would necessitate direct integration with industry-standard Computer-Aided Design (CAD) platforms.

\section*{Acknowledgments}
\vspace{-8pt}
This research was conducted using computational resources and services at the Center for Computation and Visualization, Brown University.

\section*{Funding}
\vspace{-8pt}
VK and GEK acknowledge support from Defense Advanced Research Projects Agency (DARPA) under the Automated Prediction Aided by Quantized Simulators (APAQuS) program, Grant No. HR00112490526, AFOSR Multidisciplinary Research Program of the University Research Initiative (MURI) grant FA9550-20-1-0358, ONR Vannevar Bush Faculty Fellowship (N00014-22-1-2795), and U.S. Department of Energy project SEA-CROGS (DE-SC0023191).


\clearpage
\newpage
\makeatletter
\renewcommand \thesection{A\@arabic\c@section}
\renewcommand\thetable{A\@arabic\c@table}
\renewcommand \thefigure{A\@arabic\c@figure}
\makeatother

\section*{\Large{Appendices}}

\setcounter{figure}{0}
\setcounter{table}{0}
\setcounter{section}{0}
\setcounter{page}{1}

\section{Related works}
\label{sec:related_works}
In this section, we provide an overview of existing works in three distinct but convergent research domains: Set-Based Design (SBD) for managing design space exploration, Risk-Based Design (RBD) for handling uncertainty, and the emerging field of Multi-Agent Systems (MAS) used in the context of engineering design. 

As an alternative to traditional point-based design approach, SBD involves reasoning about and manipulating sets of designs \cite{set_based_ward}. Since its introduction, SBD has been applied to various engineering problems. Canbaz et al. \cite{Framework_set_based_design} developed a collaborative SBD framework applied to a cantilever beam problem, demonstrating how sets can be managed and communicated in a distributed design environment. Hannapel, Vlahopoulos, and Singer \cite{set_based_multidisciplinary} have extensively explored the intersection between SBD and Multidisciplinary Design Optimization (MDO), first proposing principles for including SBD in MDO and later detailing a formal implementation, demonstrating how the two paradigms can be used synergistically. Riaz et al. \cite{set_based_Passenger_aircraft} applied a set-based approach to passenger aircraft family design, showcasing its utility in managing the immense complexity and inter-dependencies inherent in such systems. Similarly, Small et al. \cite{Set_based_UAV} provided a detailed case study of SBD for a UAV, highlighting its benefits in early-stage conceptual design trade-offs. Specking et al. \cite{set_based_specking} integrated SBD with Model-Based Systems Engineering for design space exploration and applied it to a UAV case study. Wade et al. \cite{set_based_Wade} utilized SBD for designing engineered resilient systems, while McKenney et al. \cite{set_based_McKenney} demonstrated its effectiveness in adapting to evolving design requirements, a key advantage in long-duration projects. To enhance the rigor of set reduction, Georgiades et al. \cite{set_based_adopt} proposed ADOPT, an augmented framework that integrates formal optimization techniques directly into the SBD workflow.

While SBD provides a framework for exploration, designing modern engineering systems requires a formal methodology for handling uncertainty. Traditional deterministic design optimizes for a single operating point, while reliability-based design optimization (RBDO) targets a specific probability of failure. Risk-Based Design (RBD) extends these concepts by incorporating the consequences of failure, defining risk as a function of both failure probability and its associated cost or severity. Beck and Gomes \cite{BECK201218} provided a comparative analysis, showing that risk-based optimization yields different and often more robust designs than deterministic or purely reliability-based methods. A foundational element of modern RBD is the use of coherent risk measures, such as the Conditional Value-at-Risk (CVaR) \cite{cvar_optimization}, (also known as superquantile) which quantifies the expected loss in the tail of a distribution. Rockafellar and Royset \cite{risk_design_Rockafellar} established the theoretical underpinnings for using such measures in engineering design, providing a mathematically sound basis for risk-averse optimization. Royset et al. \cite{risk_design_hydrofoil} proposed a Risk-Adaptive Set-Based Design framework, applying it to the shaping of a hydrofoil using CVaR as the risk metric for selecting design candidates. Chaudhuri et al. \cite{risk_design_CVaR_Anirban} investigated the use of various risk metrics, including Probability of Failure, CVaR, and Buffered Probability of Failure, and later developed methods for certifiable risk-based optimization \cite{risk_design_certifiable_Anirban}, which aims to provide high-confidence guarantees on performance. The challenge of incorporating different types of uncertainty aleatory (inherent randomness) and epistemic (lack of knowledge) has been addressed by researchers like Rumpfkeil \cite{robust_design_Rumpfkeil}, who developed robust design methods for mixed uncertainties. Li et al. \cite{risk_design_CVAR_Li} developed a CVaR-based approach for uncertain MDO and later extended it to handle hybrid uncertainties \cite{risk_design_hybrid_Li}. Application-focused studies, such as the multi-objective robust design of airfoils by Padovan et al. \cite{risk_design_Padovan}, demonstrate the practical value of these methods in achieving designs that are insensitive to variations in operating conditions.

A new research direction in engineering design automation has been catalyzed by the advent of Large Language Models (LLMs). As a step further, integrating these LLMs into a MAS framework has now been shown to solve complex tasks \cite{AI_coscientist_google, virtuallab_swanson, sciagents}. Several frameworks have been proposed to structure this collaboration for design tasks. Obieke et al. \cite{obieke2025framework} introduced AICED, a general framework for AI collaboration in design, while Ding et al.'s DesignGPT \cite{ding2023designgpt} focuses specifically on multi-agent collaboration dynamics. Zhang et al. \cite{zhang2025idesigngpt} demonstrated that agentic workflows in their iDesignGPT system can significantly boost engineering design productivity by automating complex, multi-step tasks. Application of MAS is rapidly evolving, and application examples can be seen across different design stages \cite{massoudi2025agentic, ghasemi2025vision, panta2025meda, elrefaie2025ai, concept_to_manufacture}. An important aspect of practical implementation of MAS is the the management of interaction between the agents and external software/tools that exist in practice. This integration with external scientific computing tools is a recurring theme, with frameworks like My-CrunchGPT by Kumar et al. \cite{mycrunchgpt} aiming to create LLM-assisted platforms for broader scientific machine learning, which is directly applicable to engineering analysis. In a recent work, Kumar et al. \cite{autonomous_design_PartI} demonstrate the use of knowledge-guided MAS system for design, evaluation, and design modification of NACA airfoils, with a human-in-loop approach to ensure human control over the final solution candidate. 

This research landscape presents a research opportunity: to develop a hybrid intelligence framework that leverages the strength of each domain to create a holistic design workflow. Specifically, there is a need for a multi-agent system that can orchestrate a design process grounded in Set-Based Design principles while automating some redundant tasks during concept evaluation and modification. Such a system could use LLM agents equipped with the tools necessary for the design workflow to perform the necessary functions in SBD and documenting the rationale for design selection. This approach can eventually automate the design space exploration process while ensuring that decisions are based on risk-informed quantitative evidence. 

\section{Bayesian surrogates}
\label{sec:bayesian_surrogate}
Unlike traditional neural networks that learn a single optimal value for each weight, a BNN learns a posterior distribution over its weights and biases, $p(\mathbf{w}|\mathcal{D})$, conditioned on the training data $\mathcal{D}$. As this true posterior is analytically intractable for deep neural networks, we employ variational inference (VI) to approximate it with a more manageable distribution, $q(\mathbf{w}|\theta)$, parameterized by variational parameters $\theta$ [1]. We place a standard normal distribution as a non-informative prior over the weights, $p(\mathbf{w}) = \mathcal{N}(0, \mathbf{I})$, which acts as a form of regularization. The variational posterior is modeled using a mean-field approximation, where each weight has an independent Gaussian distribution, $q(w_i|\theta_i) = \mathcal{N}(w_i|\mu_i, \sigma_i^2)$, with $\theta_i = \{\mu_i, \sigma_i\}$ being the learnable mean and standard deviation for that weight. The training objective is to find the optimal parameters $\theta^*$ by maximizing the Evidence Lower Bound (ELBO), which is equivalent to minimizing the Kullback-Leibler (KL) divergence between the approximate and true posteriors:

\begin{equation}
    \mathcal{L}(\theta) = \mathbb{E}_{q(\mathbf{w}|\theta)}[\log p(\mathcal{D}|\mathbf{w})] - \text{KL}[q(\mathbf{w}|\theta) \|\| p(\mathbf{w})]
\end{equation}

The network architecture itself is a fully-connected feed-forward model utilizing three `DenseFlipout' layers, using a Leaky ReLU activation function with a negative slope of 0.2. The output of the network consists of two neurons, which directly parameterize the predictive distribution of the target coefficient. A softplus activation function is applied to the second output neuron to ensure the predicted standard deviation is always positive. For a given input vector $\mathbf{x}$, the network outputs the mean $\mu(\mathbf{x})$ and standard deviation $\sigma(\mathbf{x})$ of a Gaussian distribution, thereby capturing both epistemic uncertainty (through the distribution over weights) and aleatoric uncertainty (inherent noise in the data). The final predictive distribution for a new data point $\mathbf{x}^*$ is obtained by marginalizing over the posterior of the weights, $p(y^*|\mathbf{x}^*, \mathcal{D}) \approx \int p(y^*|\mathbf{x}^*, \mathbf{w})q(\mathbf{w}|\theta^*)d\mathbf{w}$. For prediction, epistemic uncertainty is quantified through Monte Carlo sampling, where 200 forward passes are performed for each input, each time with a new set of weights drawn from their learned posterior distributions. The final deterministic prediction is taken as the mean of these samples, while their variance serves as a measure of the model's confidence.

\subsection{Training data}
For training our Bayesian surrogate, we utilize the CFD data published earlier in \cite{airfoil_data}. These simulations solve the Reynolds-Averaged Navier-Stokes (RANS) equations, coupled with the Spalart-Allmaras one-equation turbulence model to ensure accurate aerodynamic predictions. The dataset encompasses a total of 597 unique airfoil designs, each defined by a distinct combination of Class Shape Transformation (CST) parameters and operating conditions, specifically Mach number and angle of attack, which are varied within the bounds specified in Table \ref{tab:cst_space}. For the purpose of model training and validation, this dataset is partitioned into a pre-defined training set of 497 samples and a test set of 100 samples. Each sample in the dataset contains the integral aerodynamic performance coefficients: drag ($C_D$), lift ($C_L$), and moment ($C_M$), as well as the detailed surface pressure distribution, represented by the pressure coefficient ($C_P$) at numerous points along the airfoil chord. The Bayesian surrogate model is constructed to establish a probabilistic mapping from the input space of design and operational parameters, $\mathbf{x} \in \mathcal{X} \subset \mathbb{R}^d$, to the output space of aerodynamic coefficients, $\mathbf{y} \in \mathcal{Y} \subset \mathbb{R}$. Here, the input vector $\mathbf{x}$ comprises the CST parameters and operating conditions, while the output is one of the scalar quantities $\mathbf{y} = \{C_D, C_L, C_M\}$. For the specific design space explored in this work, this purpose-built surrogate offers a higher level of fidelity and localized accuracy compared to more generalized, pre-trained models such as Neuralfoil.

\begin{table}[]
\small
\centering
\caption{Airfoil input parameter space showing the range of CST weights $w$ and operating conditions used for our surrogate model training \cite{airfoil_data}.}
\label{tab:cst_space}
\resizebox{0.5\textwidth}{!}{%
\begin{tabular}{@{}lll@{}}
\toprule
\textbf{Parameter} & \textbf{Lower bound}    & \textbf{Upper bound}      \\ \midrule
Ma                 & 0.2                     & 0.7                       \\
AoA                & -3.0                      & 5.0                         \\
Re                 & $1\times10^6$          & $6.5 \times 10^6$             \\
$w_{u,1}$             & 0.0644                  & 0.1932                    \\
$w_{u,2}$             & 0.0688                  & 0.2064                    \\
$w_{u,3}$             & 0.0961                  & 0.2883                    \\
$w_{u,4}$             & 0.0961                  & 0.2882                    \\
$w_{u,5}$             & 0.101                   & 0.303                     \\
$w_{l,2}$             & 0.068                   & 0.2039                    \\
$w_{l,3}$             & 0.1126                  & 0.3377                    \\
$w_{l,4}$             & 0.0381                  & 0.1143                    \\
$w_{l,5}$             & -0.0586                 & -0.0195                   \\ \bottomrule
\end{tabular}%
}
\end{table}

\subsection{Training results}
\begin{figure}[h]
    \centering
    \includegraphics[width=1\linewidth]{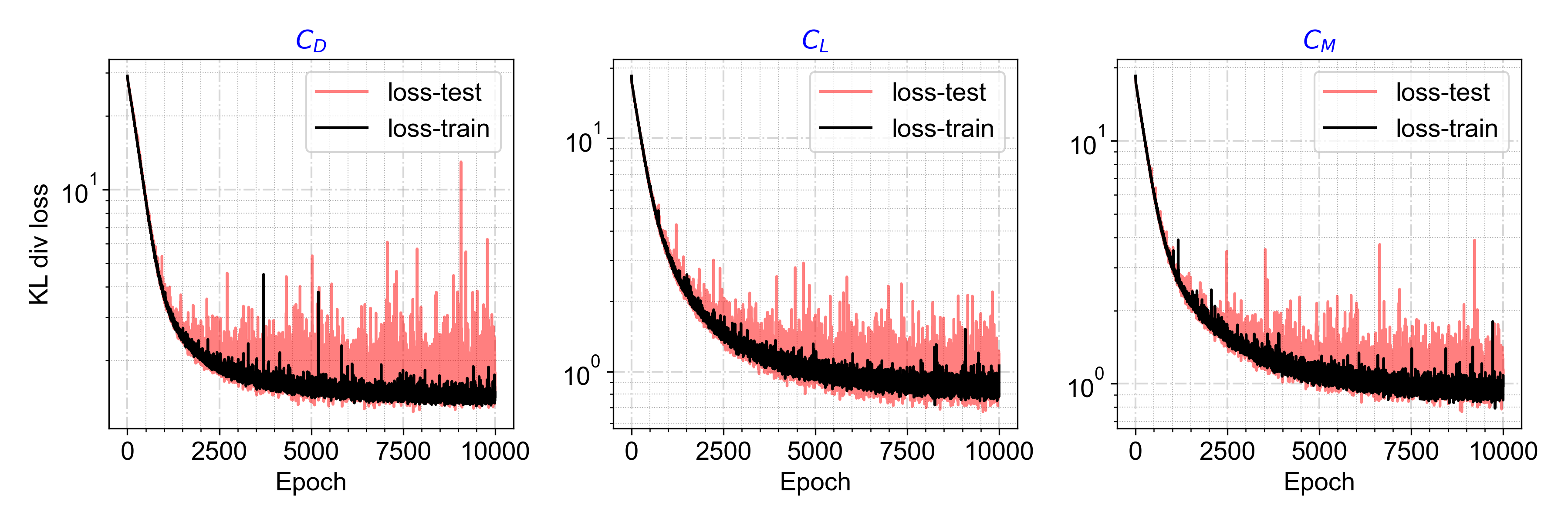}
    \caption{Loss evolution for the three Bayesian surrogates: $C_D, C_L, C_M$. }
    \label{fig:bayesian_loss}
\end{figure}

\begin{figure}[h]
    \centering
    \includegraphics[width=1\linewidth]{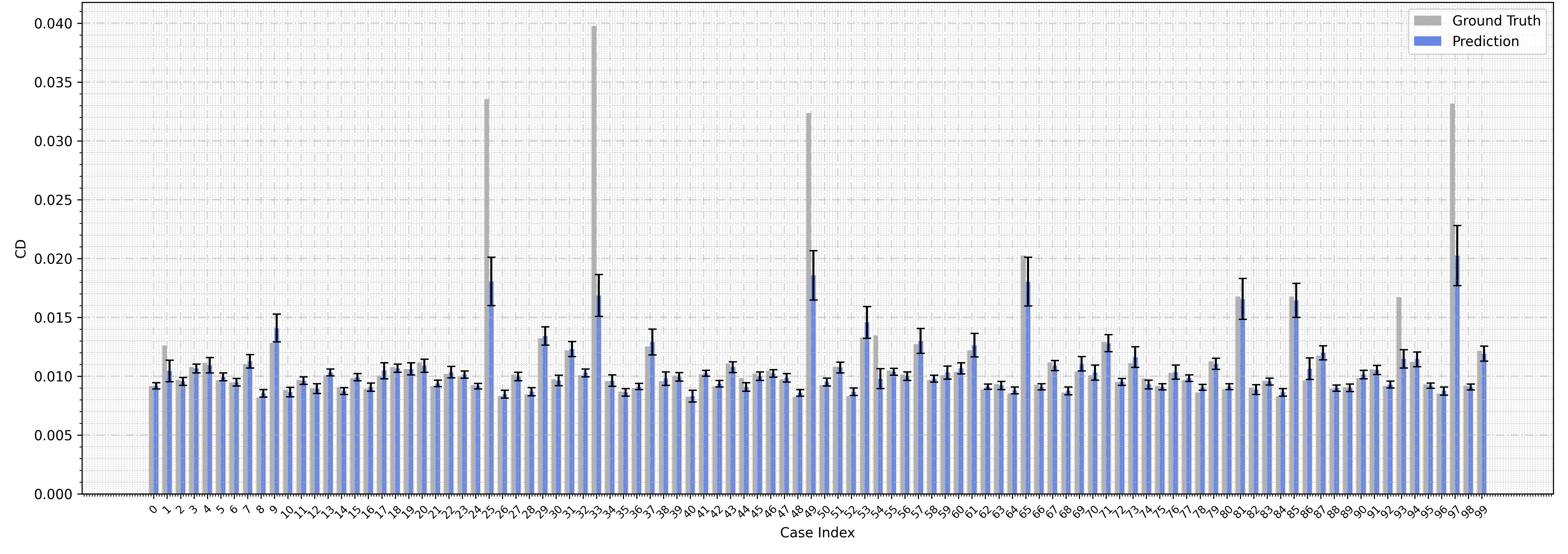}
    \caption{Ensemble prediction results from Bayesian surrogate for coefficient $C_D$. The black error bar shows the expected uncertainty in prediction combining both aleatoric and epistemic uncertainties. }
    \label{fig:bayesian_pred_cd}
\end{figure}

\begin{figure}[h]
    \centering
    \includegraphics[width=1\linewidth]{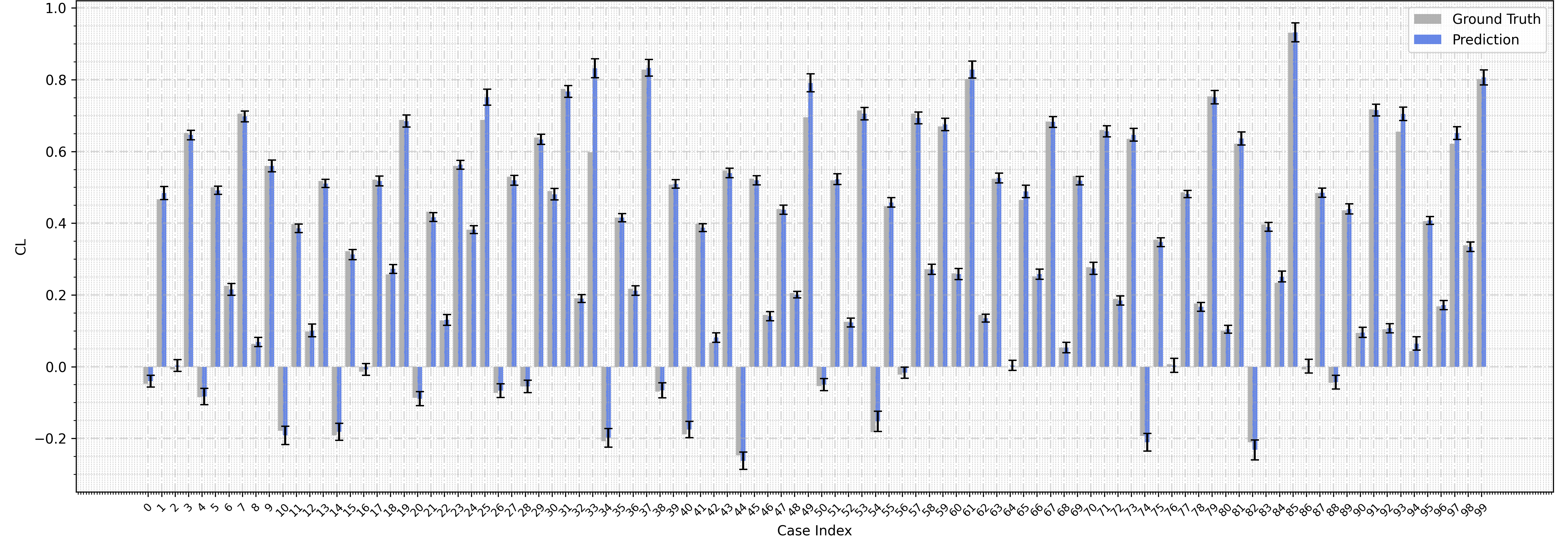}
    \caption{Ensemble prediction results from Bayesian surrogate for coefficient $C_L$. The black error bar shows the expected uncertainty in prediction combining both aleatoric and epistemic uncertainties. }
    \label{fig:bayesian_pred_cl}
\end{figure}

\begin{figure}[h]
    \centering
    \includegraphics[width=1\linewidth]{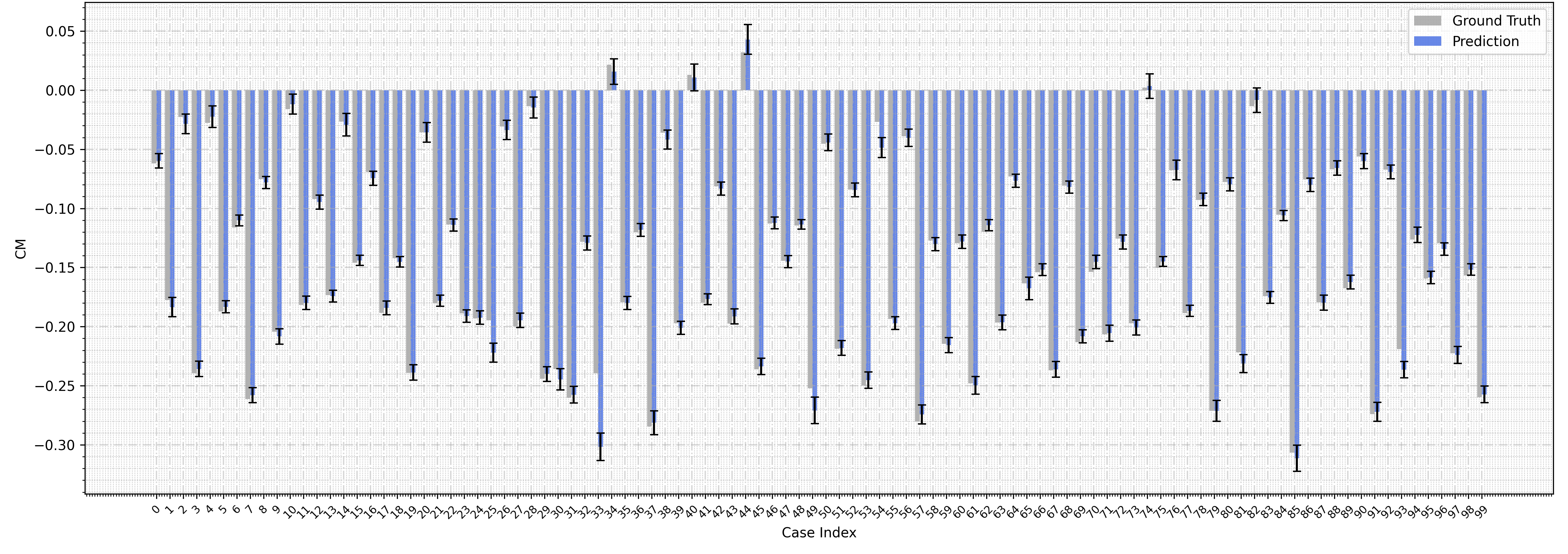}
    \caption{Ensemble prediction results from Bayesian surrogate for coefficient $C_M$. The black error bar shows the expected uncertainty in prediction combining both aleatoric and epistemic uncertainties.}
    \label{fig:bayesian_pred_cm}
\end{figure}

The loss convergence plot for the three Bayesian networks trained separately is shown in figure \ref{fig:bayesian_loss}. All models converge to a KL divergence loss value of $\approx 1$ over 10,000 epochs. A prediction distribution for each test case is obtained by sampling the Bayesian models 200 times, resulting in a distribution of coefficient predictions for each airfoil design and operating conditions. In figure \ref{fig:bayesian_pred_cd} - \ref{fig:bayesian_pred_cm}, we show the ensemble prediction results for the three coefficients across the test set. From this plot, we note that the Bayesian surrogates can accurately predict the aerodynamic coefficients for different airfoil designs and operating conditions. The $C_D$ values for some samples, such as sample 31 shows significant departure from the average values that occur across different test samples. Some samples, in figure \ref{fig:bayesian_pred_cl} also show negative lift, which occurs since the randomly generated airfoil designs may not always be aerodynamically consistent with expectations.

\section{DeepONet surrogate}
\label{sec:DeepONet}
The DeepONet surrogate consists of two main components: a `branch' network that encodes the input function parameters and a `trunk' network that processes the spatial coordinates where the output function is evaluated. In our formulation, we define an operator $\mathcal{G}$ that maps the combined geometry and flow conditions to the pressure coefficient function, $C_P$. The input to the operator is a vector $\mathbf{u} \in \mathbb{R}^{12}$, which concatenates the vector of Class Shape Transformation (CST) weights, $\mathbf{w}$, with the flow conditions: Mach number ($Ma$), angle of attack ($AoA$), and Reynolds number ($Re$). The output of the operator, $\mathcal{G}(\mathbf{u})$, is a function that maps the airfoil surface coordinates, $\mathbf{x} \in \mathbb{R}^2$, to the scalar pressure coefficient, $C_P(\mathbf{x}, \mathbf{u}) \in \mathbb{R}$. The DeepONet approximates this operator as $\mathcal{G}(\mathbf{u})(\mathbf{x}) \approx \mathbf{b}(\mathbf{u})^T \mathbf{t}(\mathbf{x})$, where $\mathbf{b}: \mathbb{R}^{12} \to \mathbb{R}^{p}$ represents the branch network and $\mathbf{t}: \mathbb{R}^{2} \to \mathbb{R}^{p}$ is the trunk network, with both mapping to a shared latent space of dimension $p$. The training data, comprising surface coordinates and corresponding $C_P$ values, was sourced from the same high-fidelity CFD dataset used for the Bayesian surrogates. Both the branch and trunk networks are implemented as Multi-Layer Perceptrons (MLPs) with five hidden layers configured with [64, 64, 128, 128, 128] units and utilize the Swish activation function throughout \cite{swish_activation}. The latent dimension is set to $p=128$. For preprocessing, branch inputs and target $C_P$ values are normalized via min-max scaling, while trunk inputs are scaled to the range [-1, 1]. To enhance generalization, the branch network incorporates dropout with a rate of 0.05, while the trunk network employs Layer Normalization between successive layers. The model is trained by minimizing the Mean Squared Error (MSE) loss function using the Lion optimizer \cite{Lion_optimizer}, with a batch size of 100. The learning rate is managed by an exponential decay schedule, starting at $1 \times 10^{-3}$ and decaying with a rate of 0.96 every 500 steps. Model performance on unseen designs is assessed using the relative $L^2$ error norm on the test set.

\subsection{Training results}
\begin{figure}[H]
    \centering
    \includegraphics[width=0.65\linewidth]{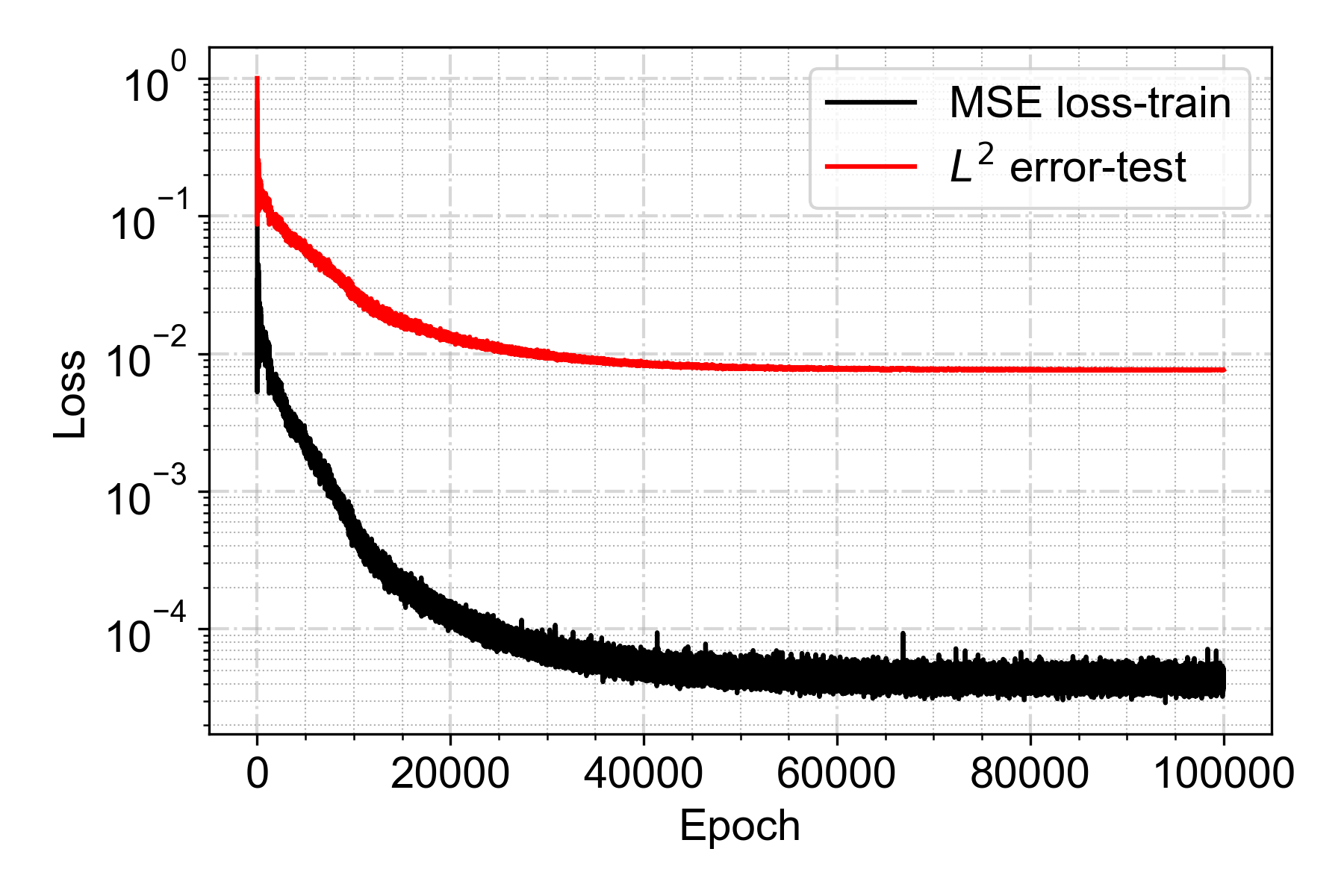}
    \caption{Loss evolution for the DeepONet showing convergence. }
    \label{fig:deeponet_loss}
\end{figure}

\begin{figure}[H]
    \centering
    \includegraphics[width=1\linewidth]{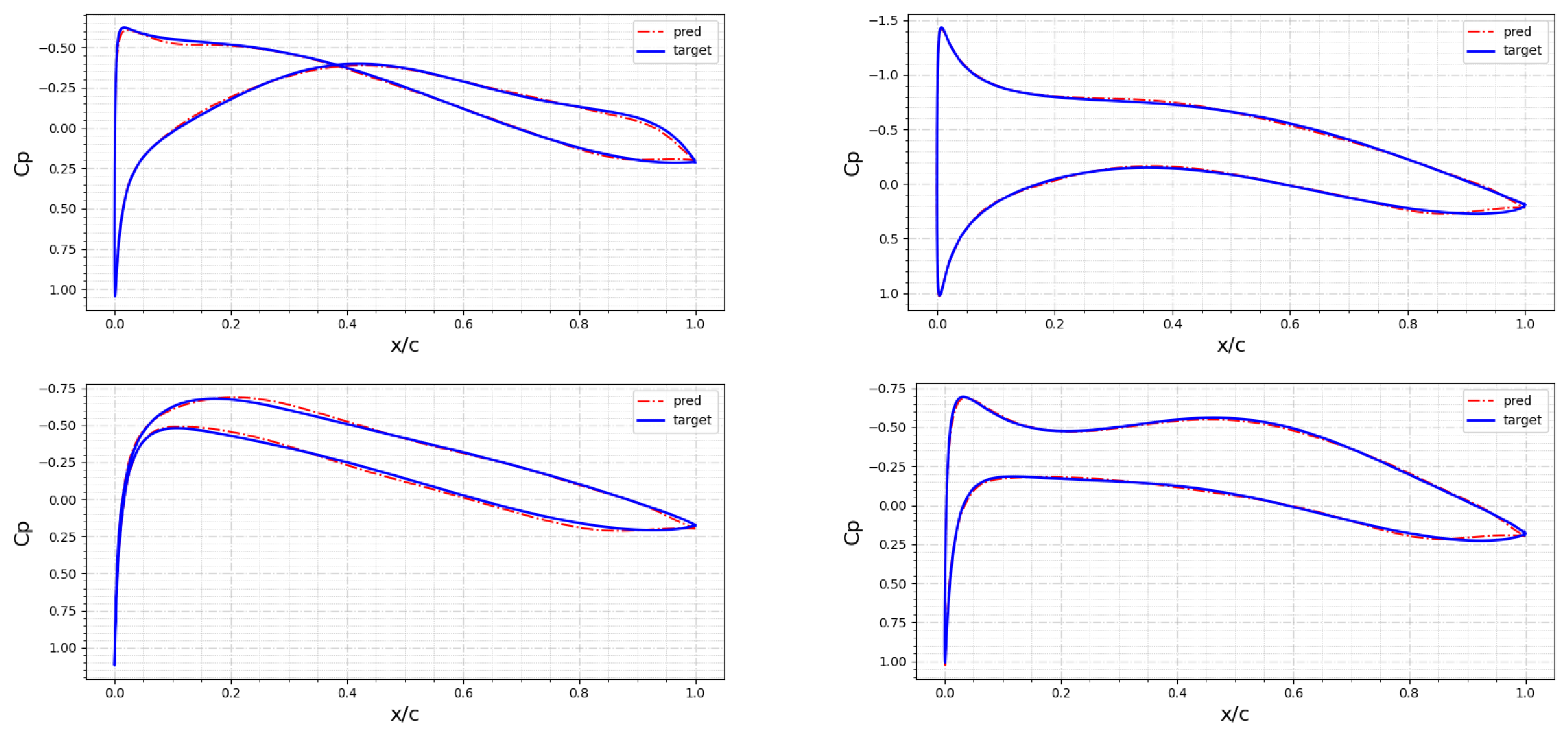}
    \caption{Samples of DeepONet prediction for coefficient of pressure, $C_P$.}
    \label{fig:deeponet_prediction}
\end{figure}

Figure \ref{fig:deeponet_loss} shows the loss convergence for our DeepONet surrogate. The median $L^2$ relative error, calculated individually across all 100 samples in the test set was found to be $\approx 5.9\%$. While the prediction was generally acceptable for most test cases, a few instances were observed where the prediction error was higher than the median error. However, the overall prediction from our DeepONet surrogate on this dataset was within acceptable threshold for this study. In figure \ref{fig:deeponet_prediction}, we show the prediction results for four test cases with our surrogate, where a reasonable fit to ground truth data is observed.

\section{Stage 3: Sensitivity Analysis}
In this section, we show the two plots that were generated during Stage 3: Sensitivity analysis of the 9 CST parameters and how they affect the aerodynamic performance coefficients. These results were generated by the Analyst agent, which re-uses these plots along with numerical values of the Sobol indices to determine the nature of response expected by increasing or decreasing a CST parameter. 

\begin{figure}[H]
    \centering
    \includegraphics[width=1\linewidth]{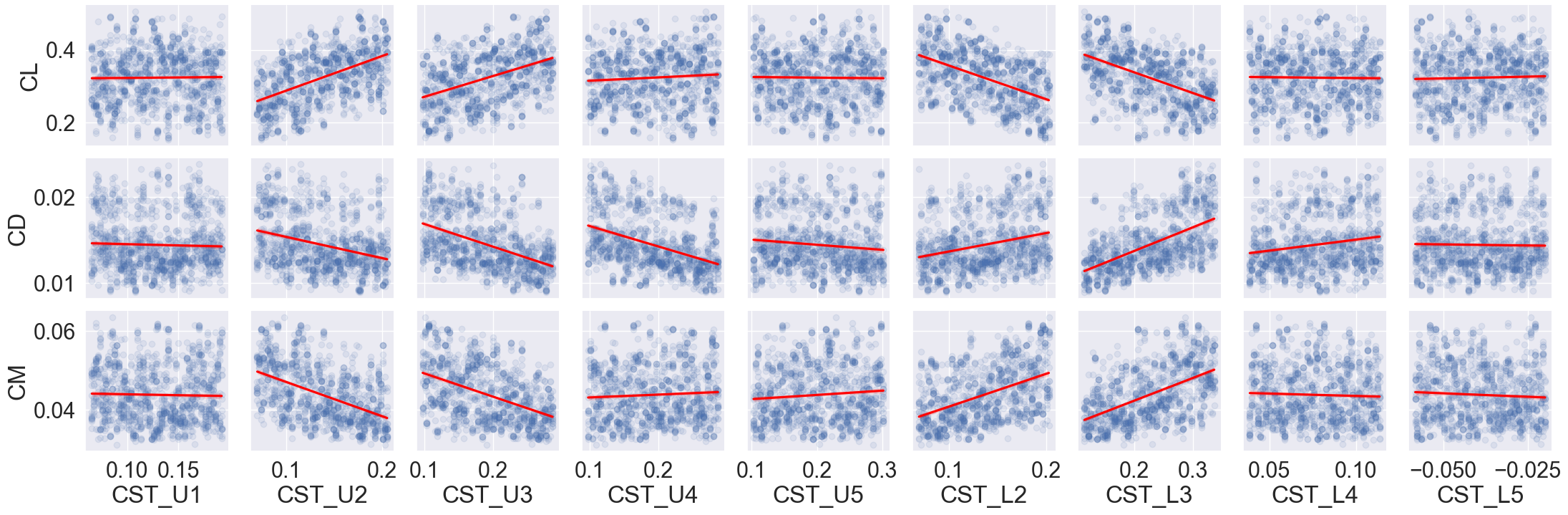}
    \caption{Pairwise plot showing the individual effect of changing CST parameters on aerodynamic coefficients $C_D$, $C_L$, and $C_M$.}
    \label{fig:param_effect_pairwise}
\end{figure}

\begin{figure}[H]
    \centering
    \includegraphics[width=0.7\linewidth]{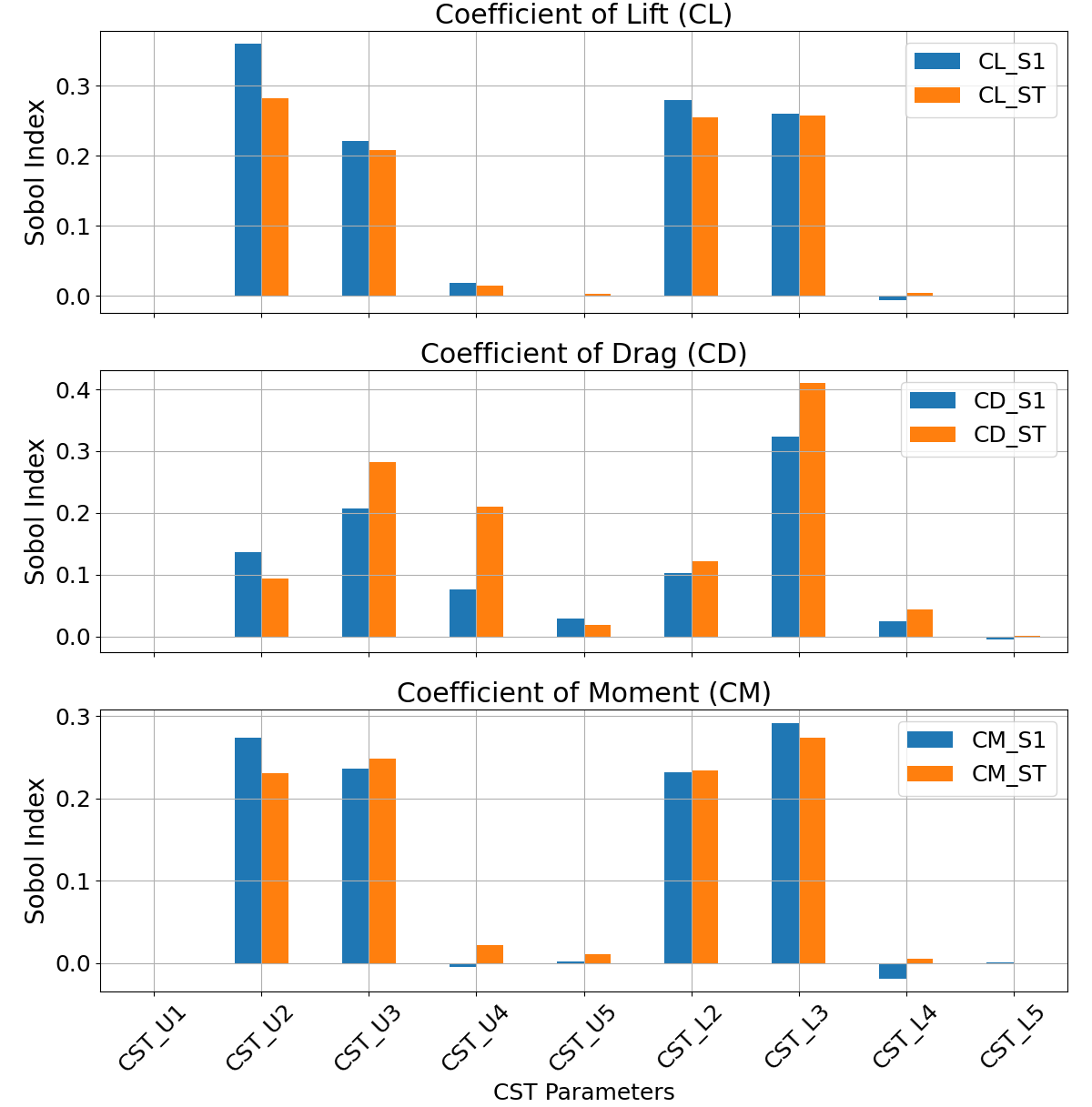}
    \caption{Plot showing magnitude of Sobol indices showing individual CST parameter effect ($S_i$) and the total effect ($S_{Ti}$) for each aerodynamic coefficient.}
    \label{fig:sobol_indx}
\end{figure}

\newpage
\section{Stage 6: Automated Design review (additional information)}
In this section, we provide additional information related to the automated design review phase performed by the Systems Engineer agent.

\subsection{RAE2822 benchmark}
\begin{figure}[H]
    \centering
    \includegraphics[width=0.9\linewidth]{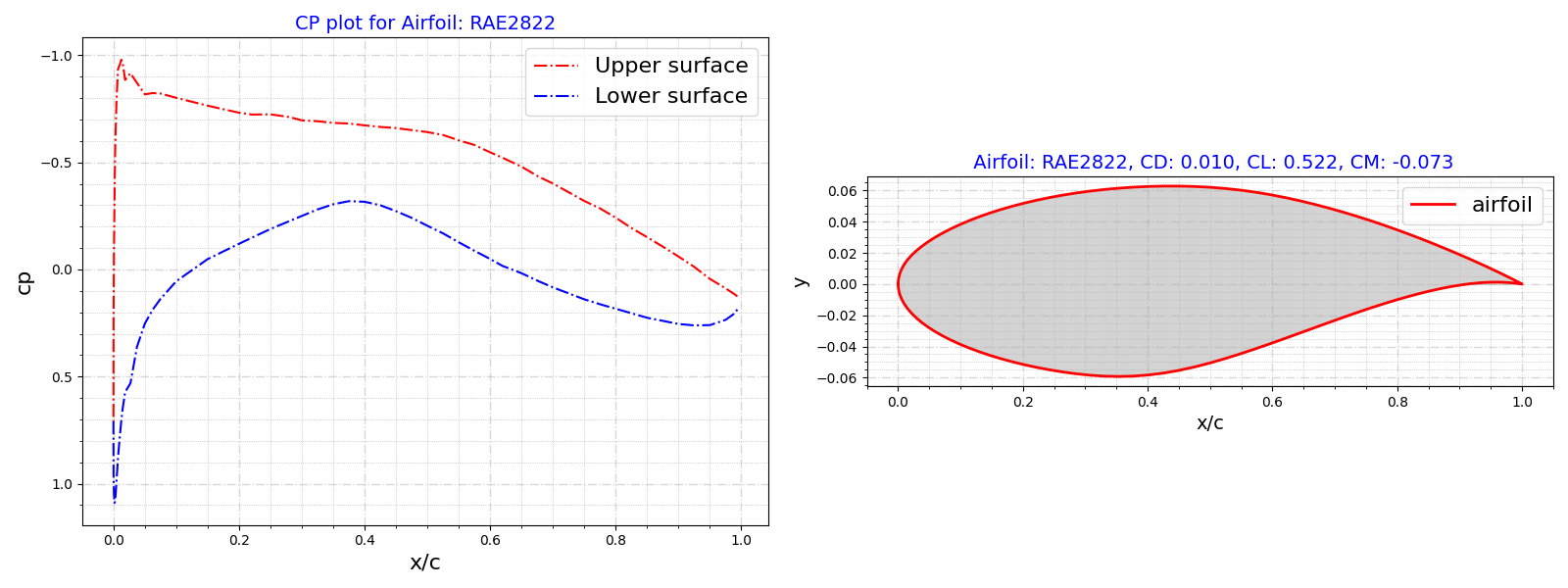}
    \caption{Plot showing the coefficient of pressure plot for benchmark airfoil RAE2822 at Re=6.3 million, Ma=0.6, and AoA=2.57 degrees \cite{rae2822_report}. The airfoil profile is shown alongside. The Systems Engineer agent compares the $C_P$ plot along with utility score for the new airfoil designs against this benchmark and determines whether the design should be considered valid or not based on a set of heuristic rules defined by the human Manager.}
    \label{fig:RAE2822}
\end{figure}

\subsection{Engineering feedback generated during iterative design review and update loop}
\begin{figure}[H]
    \centering
    \begin{response}{Design feedback for Design ID-486 - iteration 3}
    \scriptsize
        ``\textbf{Pressure Distribution Curve Assessment}:
        \begin{enumerate}
            \vspace{-5pt}\item The airfoil ID-486 exhibits a pronounced suction peak on the upper surface near the leading edge (around x/c = 0.05-0.1), which contributes to a high lift coefficient but could potentially lead to early shock formation or flow separation at higher Mach numbers.
            \vspace{-5pt} \item The pressure recovery on the upper surface is relatively smooth, indicating good flow behavior towards the trailing edge.
            \vspace{-5pt} \item Compared to the benchmark RAE2822, ID-486 generates a higher pressure differential between the upper and lower surfaces, consistent with its higher CL. However, the benchmark's pressure distribution appears more tailored for supercritical flow with a more gradual suction peak.
        \end{enumerate}
        \textbf{Airfoil Profile Assessment}:
        \begin{enumerate}
            \vspace{-5pt}\item The airfoil ID-486 has a generally smooth and continuous profile, which is favorable for manufacturability and structural integrity.
            \vspace{-5pt} \item The airfoil appears to have a reasonable thickness distribution, providing adequate volume for internal structures and maintaining structural integrity.
            \vspace{-5pt} \item The overall shape is comparable to the benchmark RAE2822 in terms of non-functional requirements, with no apparent features that would hinder maintenance or service.
        \end{enumerate}
        \textbf{Design improvements}:
        \begin{enumerate}
            \vspace{-5pt} \item Increase CST\_U2: This parameter has a highly beneficial effect, as increasing it will simultaneously increase the Coefficient of Lift (CL), decrease the Coefficient of Drag (CD), and increase the Coefficient of Moment (CM), making the pitching moment less negative and improving stability.
            \vspace{-5pt} \item Decrease CST\_U3: Similar to CST\_U2, decreasing CST\_U3 will also lead to an increase in CL, a decrease in CD, and an increase in CM, contributing positively across all three aerodynamic coefficients.
            \vspace{-5pt} \item Refine Leading Edge Curvature: To potentially mitigate the strong suction peak and further optimize drag, adjustments to the leading edge curvature, possibly through fine-tuning CST\_U1 or further adjustments to CST\_U2 and CST\_U3, could be explored to achieve a smoother pressure distribution while maintaining high lift.
        \end{enumerate}
        Pressure rating: 4; Valid: True ''
        
        \centering
        \includegraphics[width=0.6\linewidth]{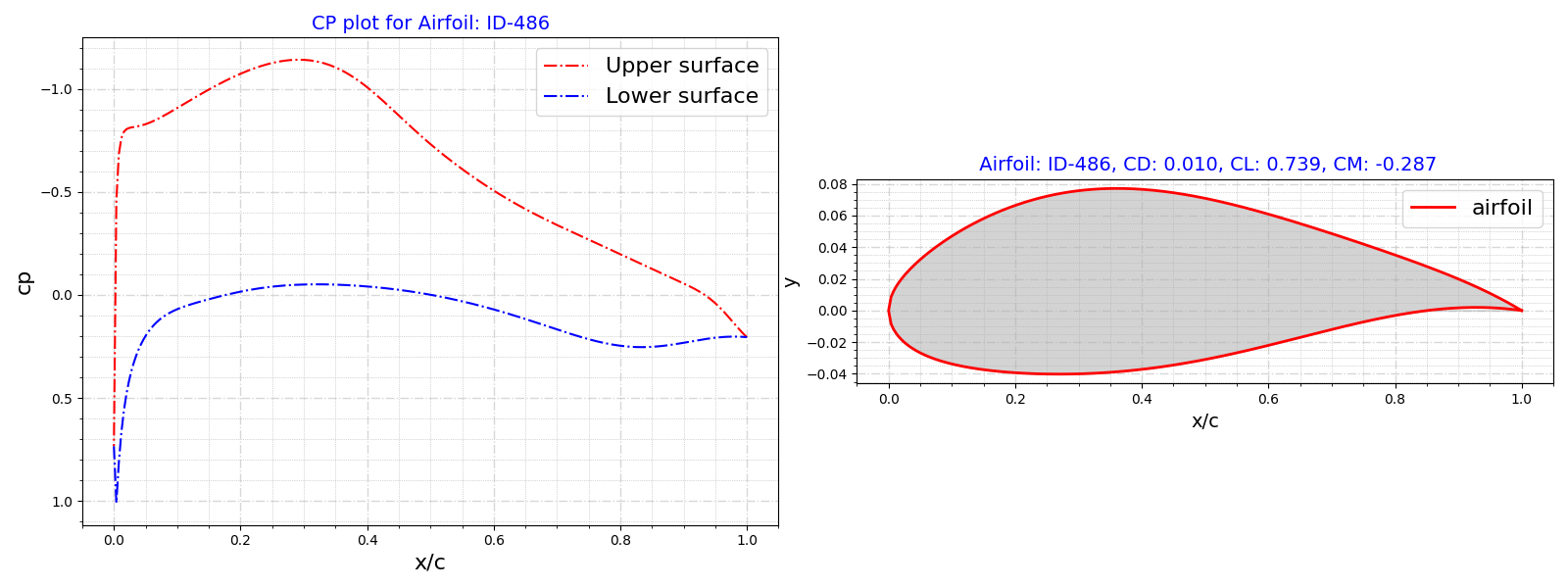}
    \end{response}
    \caption{Feedback for design ID-486 generated by the Systems Engineer with human Manager during third iteration after design modification.}
    \label{fig:design_feedback_iter3}
\end{figure}

\begin{figure}[H]
    \centering
    \begin{response}{Design feedback for Design ID-486 - iteration 4}
    \scriptsize
        ``\textbf{Pressure Distribution Curve}:
        \vspace{-5pt}
        \begin{enumerate}
            \vspace{-5pt} \item Upper Surface Suction Peak: The airfoil ID-486 exhibits a strong suction peak near the leading edge (x/c ~0.05) on the upper surface, which is more pronounced than the benchmark RAE2822. This indicates higher local velocities and contributes significantly to the higher Coefficient of Lift (CL) observed for ID-486.
            \vspace{-5pt} \item Pressure Recovery: Both the upper and lower surfaces show a smooth and gradual pressure recovery towards the trailing edge, suggesting that there is no significant flow separation, which is a positive indicator for aerodynamic efficiency.
            \vspace{-5pt} \item Lift Generation: The larger pressure differential between the upper and lower surfaces, particularly due to the deeper suction on the upper surface, confirms the higher lift generation capability of ID-486 compared to the benchmark.
        \end{enumerate}
        \vspace{-5pt}
        \textbf{Airfoil Profile}:
        \begin{enumerate}
            \vspace{-5pt} \item Thickness and Camber: Airfoil ID-486 appears to be thicker and exhibits more camber, especially in the forward and mid-chord regions, compared to the benchmark RAE2822. This increased camber is consistent with its higher lift performance.
            \vspace{-5pt} \item Structural Integrity: The thicker profile of ID-486 generally suggests better potential for structural integrity and allows for more internal volume for structural elements (e.g., spars) or systems, which is a favorable non-functional requirement.
            \vspace{-5pt} \item Manufacturability: The overall smooth and continuous shape of the airfoil suggests good manufacturability, similar to the benchmark, without any sharp corners or abrupt changes in curvature that could pose manufacturing challenges.
        \end{enumerate}
        \vspace{-5pt}
        \textbf{Aerodynamic Performance and Utility Scores}:
        \begin{enumerate}
           \vspace{-5pt} \item Coefficient of Lift (CL): ID-486 (CL=0.732) significantly outperforms the benchmark (CL=0.522), resulting in a much higher utility score for CL (ID-486: 0.5760 vs. Benchmark: 0.177).
           \vspace{-5pt} \item Coefficient of Drag (CD): ID-486 (CD=0.0102) has slightly higher drag than the benchmark (CD=0.010). Consequently, its utility score for CD (ID-486: 0.5147) is marginally lower than the benchmark (0.518).
           \vspace{-5pt} \item Coefficient of Moment (CM): ID-486 (CM=-0.2827) exhibits a significantly more negative pitching moment compared to the benchmark (CM=-0.073). This results in a much lower utility score for CM (ID-486: 0.0577 vs. Benchmark: 0.7566), indicating a stronger nose-down tendency which might require more trim effort.
           \vspace{-5pt} \item Combined Utility Score: Despite the lower CM utility, ID-486 achieves a higher combined utility score (0.4539) than the benchmark (0.3955), primarily driven by its superior lift performance.
        \end{enumerate}
        \vspace{-5pt}
        \textbf{Ways to Improve Performance through Design Modifications (\emph{Focusing solely on CM as per manager's feedback})}:
        \begin{enumerate}
            \vspace{-5pt} \item Increase CST\_L3: Based on the sensitivity analysis, increasing CST\_L3 is the most sensitive parameter for CM and shows a positive correlation, meaning it will increase CM (make it less negative).
            \vspace{-5pt} \item Increase CST\_U2: This parameter is highly sensitive and indicates a positive correlation with CM. Increasing CST\_U2 will help increase CM.
            \vspace{-5pt} \item Increase CST\_L2: This parameter is also highly sensitive and shows a positive correlation with CM. Increasing CST\_L2 will contribute to increasing CM.
            \vspace{-5pt} \item Decrease CST\_U3: This parameter has a high sensitivity and a negative correlation with CM. Decreasing CST\_U3 will help increase CM.
        \end{enumerate}
        Pressure rating: 4; Valid: True ''
        
        \centering
        \includegraphics[width=0.6\linewidth]{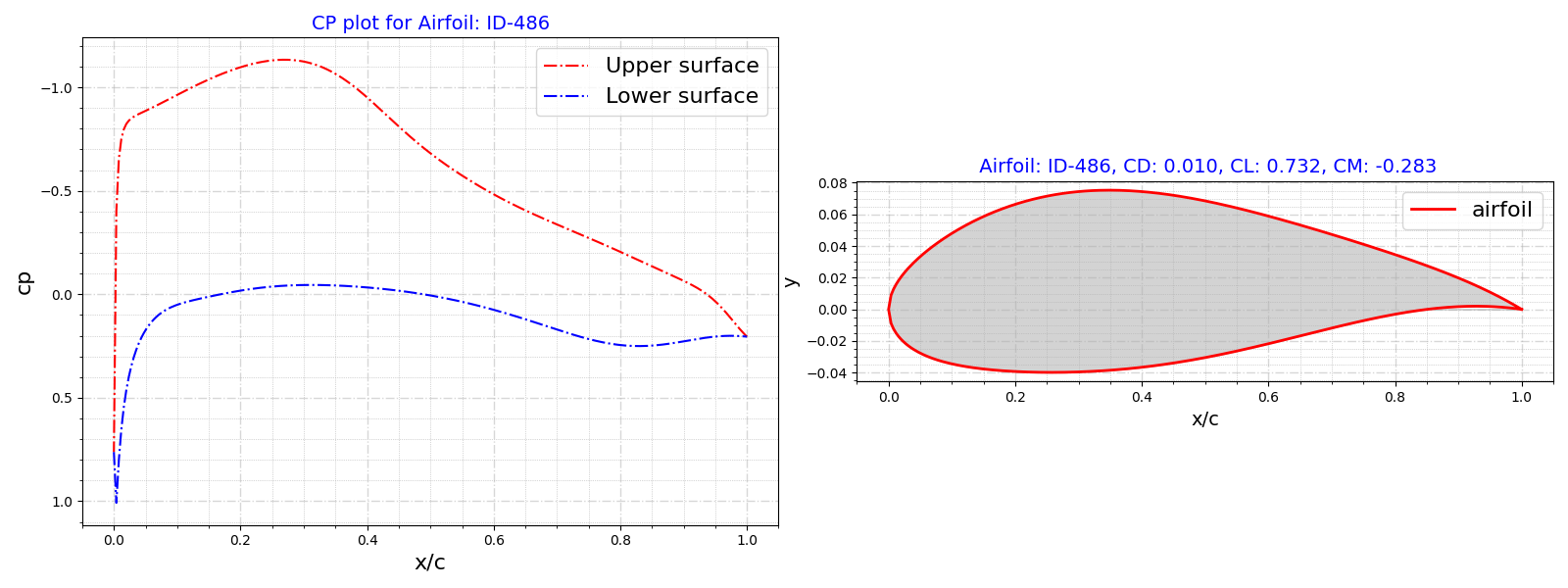}
    \end{response}
    \caption{Feedback for design ID-486 generated by the Systems Engineer with human Manager during fourth iteration after design modification.}
    \label{fig:design_feedback_iter4}
\end{figure}

\section{Stage 7: OpenFoam CFD additional results}
In this section, we provide additional CFD simulation results for the final four design candidates selected at the end of the workflow. This CFD data is used, in conjunction with coefficient of pressure and performance metrics to determine the final airfoil design/s to be developed further.
\begin{figure}[H]
    \centering
    \includegraphics[width=0.8\linewidth]{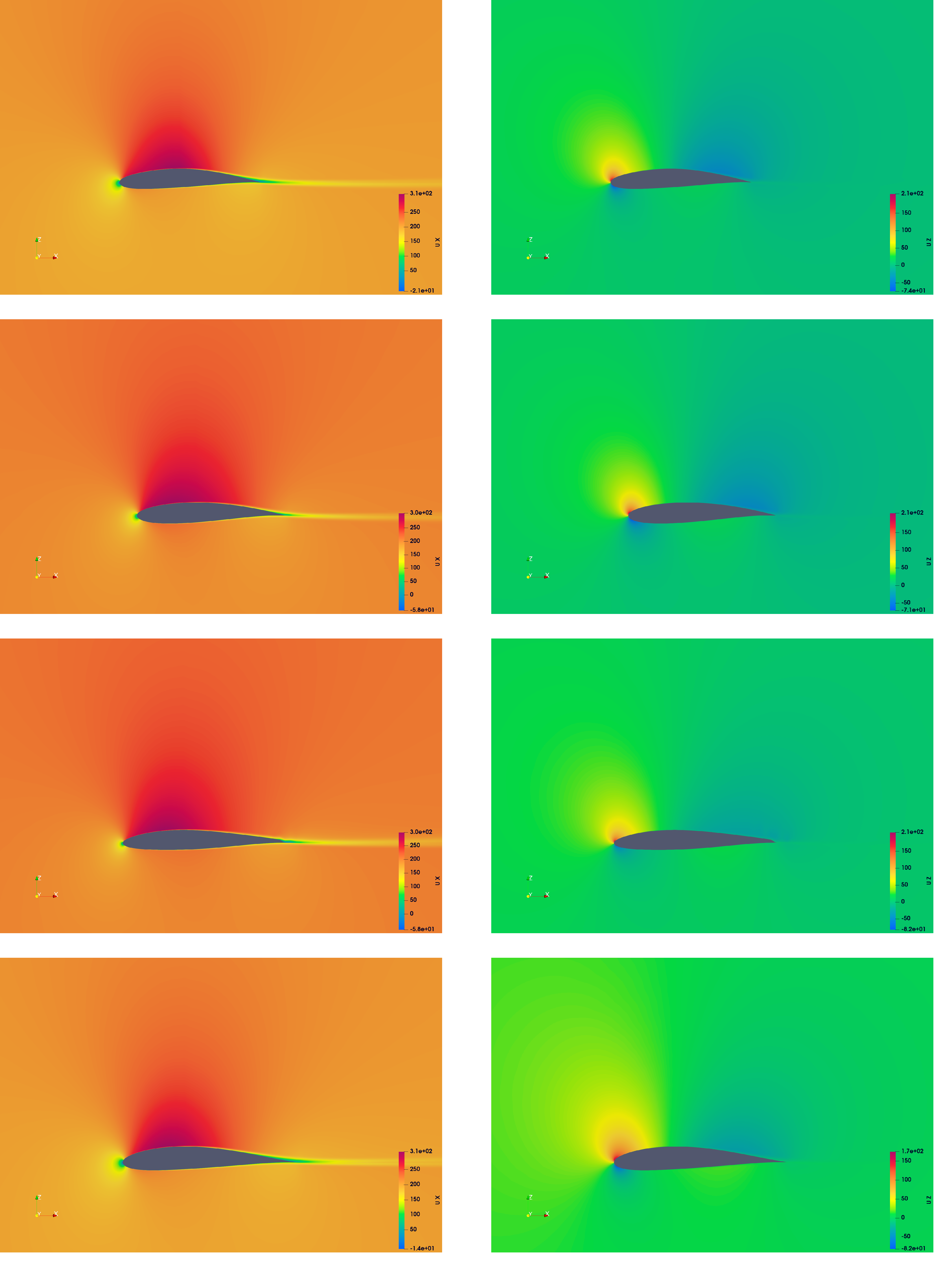}
    \caption{Plot showing velocity fields for the final design candidates selected during the agentic workflow.}
    \label{fig:cfd_velocity}
\end{figure}

\newpage
\bibliographystyle{elsarticle-num} 
\begin{footnotesize}
\bibliography{references}
\end{footnotesize}
\end{document}